\documentclass{article}

\usepackage{amsbsy,amsfonts,amsmath,natbib,sectsty,graphicx,cancel}
\usepackage{caption}
\usepackage{fncychap}
\usepackage{amssymb,subfigure,epsfig,rotating,multicol}
\usepackage{pifont,epsf,amsfonts,pstricks,pst-node,xspace,graphics,wrapfig}
\usepackage{algorithm}
\usepackage{tikz}
\usepackage{fullpage}
\usetikzlibrary{arrows,shapes,backgrounds,through,shadows,snakes}
\usetikzlibrary{calc}
\usepackage{natbib}

\newcommand{\figref}[1]{Fig. \ref{#1}}
\newcommand{\secref}[1]{Section \ref{#1}}
\renewcommand{\eqref}[1]{Eq. \ref{#1}}

\tikzstyle{cont}=[circle,draw,minimum size=6mm,line width=2pt,>=stealth]  
\tikzstyle{ocont}=[ellipse,draw=blue!50,thick,minimum size=6mm,>=stealth]  
\tikzstyle{blackcont}=[circle,draw=black!50,thick,minimum size=6mm,line width=2pt,>=stealth]  
\tikzstyle{oval}=[ellipse,draw=blue!50,thick,minimum size=6mm,line width=2pt,>=stealth]  
\tikzstyle{disc}=[rectangle,draw=blue!50,thick,line width=1pt,minimum size=6mm]  
\tikzstyle{obs}=[fill=blue!20,thick]  

\tikzstyle{fillred}=[fill=red!20,thick]  
\tikzstyle{fillgreen}=[fill=green!20,thick]  
\tikzstyle{purered}=[fill=red]  
\tikzstyle{state}=[rectangle,fill=red!20]  
\tikzstyle{sobs}=[fill=green!15,thick]  
\tikzstyle{fact}=[fill,minimum size=1.5mm,line width=2pt,>=stealth]
\tikzstyle{varfact}=[draw,minimum size=1.5mm,line width=2pt,>=stealth]
\tikzstyle{sep}=[rectangle,draw=magenta!50,thick,minimum size=6mm]  
\tikzstyle{det}=[fill=red!15,rectangle,draw=red!50,thick,minimum size=6mm]  

\tikzstyle{dethid}=[diamond,draw=red!50,thick,minimum size=6mm]  

\tikzstyle{lineball}=[fill,-*,draw=red!50,line width=1.5pt]
\tikzstyle{redball}=[mark=*,mark options={fill=red!50,draw=red},mark size=0.5pt]
\tikzstyle{greenball}=[mark=*,mark options={fill=green!50,draw=green},mark size=0.5pt]
\tikzstyle{hid}=[circle,draw,thick]  

\tikzstyle{dec}=[rectangle,draw=red!50,thick,minimum size=6mm]  
\tikzstyle{utility}=[diamond,draw=red!50,thick,minimum size=6mm]  
\tikzstyle{contdec}=[circle,draw=blue!50,thick,fill=blue!10,line width=2pt]  
\tikzstyle{decutility}=[diamond,draw=red!50,thick,minimum size=6mm]  

\tikzstyle{contobs}+=[cont]
\tikzstyle{contobs}+=[obs]
\tikzstyle{discobs}+=[disc]
\tikzstyle{discobs}+=[obs]

\tikzstyle{obsred}+=[obs]
\tikzstyle{obsred}+=[red]

\tikzstyle{background grid}=[draw, black!50,step=.1cm]
\tikzstyle{dgraph}=[->, line width=1.5pt]
\tikzstyle{ugraph}=[line width=1.5pt]

\def\ci{\perp\!\!\!\perp}

\definecolor{MyLightMagenta}{cmyk}{0.1,1,1,0.5}
\definecolor{MyGreenp}{cmyk}{0.6,0.1,0.6,0.6}
\definecolor{MyBluep}{cmyk}{0.1,1,1,0.1}

\chapterfont{\fontsize{14}{16}\selectfont}
\sectionfont{\fontsize{11}{12}\selectfont}
\subsectionfont{\fontsize{10}{11}\selectfont}
\ChNameVar{\large\sf}
\ChRuleWidth{0.5pt}
\ChNumVar{\Large}
\ChTitleVar{\Large\sf}

\newcommand{\av}[1]{\langle #1 \rangle}
\newcommand{\argmax}{\mathop{\rm arg\,max}}

\newcommand{\trans}{^{\textsf{T}}}

\newcommand{\HMSMs}{HMSMs}
\newcommand{\SARM}{SARM}
\newcommand{\GSARM}{GSARM}
\newcommand{\USARM}{USARM}
\newcommand{\HMMs}{HMMs}
\newcommand{\HMM}{HMM}

\newcommand{\SLGSSM}{SLGSSM}
\newcommand{\LGSSM}{LGSSM}

\title{Unified Treatment of Hidden Markov Switching Models}

\author{Silvia Chiappa}

\begin{document}
\maketitle

\abstract{
Many real-world problems encountered in several disciplines deal with the modeling
of time-series containing different underlying dynamical regimes, for which probabilistic
approaches are very often employed. In this paper we describe several such
approaches in the common framework of graphical models.
We give a unified overview of models previously introduced in the literature,
which is simpler and more comprehensive than previous descriptions and enables
us to highlight commonalities and differences among models that were not observed in
the past. In addition, we present several new models and inference routines,
which are naturally derived within this unified viewpoint.
}

\section{Introduction}
\vspace{-0.3cm}
Several problems encountered in application areas such as finance,
biology, speech analysis, control engineering, robotics, etc.
require the modeling of time-series containing switching among different
dynamics regimes (see \cite{ephraim02hidden} for a review).
For example, system fault diagnosis deals with detecting behavioural deviations from normality
originated by failures in the system.

Such a modeling is often achieved by employing probabilistic approaches in which
regime switching is described by a set of discrete hidden random variables, related by a
first-order Markovian dependence. All such models, that we call hidden Markov switching models (\HMSMs),
can be viewed as extensions of the popular hidden Markov model \cite{rabiner89tutorial}.

The wide interdisciplinary attention to this research area has produced
many different \HMSMs~as well as different approaches and implementations of \HMSMs~of
fundamentally similar structure, resulting in a dense literature from which extracting differences
and commonalities among models is often challenging.

In this paper we provide a simple unified treatment of existing \HMSMs, highlighting properties
and connections that were not observed in previous review papers \cite{ephraim02hidden,gales93theory,murphy2002hsm,ostendorf96from,rabiner89tutorial,yu10hidden}, and introduce novel extensions. Our exposition enables a deep understanding of the fundamental structure and relations of different approaches.
This is achieved by using the framework of graphical models, which allows to easily define complex models by using a graphical representation and to derive efficient inference routines by visual inspection of the graph, avoiding complex algebraic manipulations.

\section{Independence in Belief Networks\label{sec:GM}}
One of the fundamental tasks in probabilistic time-series modeling is the development efficient inference routines, which requires exploiting statistical independence relations among random variables\footnote{Due to space limitations, parameter learning is not discussed in this exposition. This problem however does not normally pose particular challenges once inference has been achieved.}. By using the framework of graphical models and, in particular, of belief networks \cite{barber10bayesian,kollerl09probabilistic,pearl88probabilistic}, statistical independence can be assessed by visual inspection of the graph, and therefore inference can be achieved without the need of algebraic manipulations. In this section, we introduce some basic definitions of the graphical model theory and explain two equivalent methods for assessing statistical independence. These methods form the basis for the derivations of the inference routines presented in the rest of the paper.

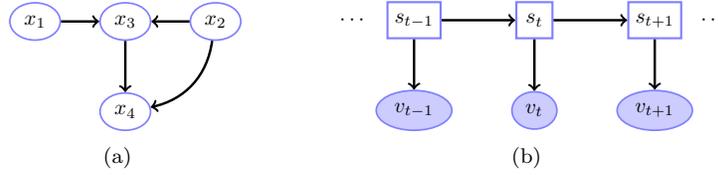
\begin{figure}[t]
\renewcommand{\subfigcapskip}{0pt}
\begin{center}
\subfigure[]{\scalebox{0.8}{
\begin{tikzpicture}[dgraph]
\node[ocont] (x2) at (4.5,0) {$x_2$};
\node[ocont] (x1) at (1.5,0) {$x_1$};
\node[ocont] (x3) at (3,0) {$x_3$};
\node[ocont] (x4) at (3,-1.5) {$x_4$};
\draw[line width=1.15pt](x1)--(x3);\draw[line width=1.15pt](x2)--(x3);\draw[line width=1.15pt](x3)--(x4);\draw[line width=1.15pt](x2)to [bend left=35](x4);
\end{tikzpicture}}}
\hskip1cm
\subfigure[]{\scalebox{0.8}{
\begin{tikzpicture}[dgraph]
\node[] at (1,0) {$\cdots$};
\node[disc] (sigmatm) at (2,0) {$s_{t-1}$};
\node[disc] (sigmat) at (4,0) {$s_t$};
\node[disc] (sigmatp) at (6,0) {$s_{t+1}$};
\node[] at (7,0) {$\cdots$};
\node[ocont,obs] (vtm) at (2,-1.5) {$v_{t-1}$};
\node[ocont,obs] (vt) at (4,-1.5) {$v_t$};
\node[ocont,obs] (vtp) at (6,-1.5) {$v_{t+1}$};
\draw[line width=1.15pt](sigmatm)--(sigmat);\draw[line width=1.15pt](sigmat)--(sigmatp);
\draw[line width=1.15pt](sigmatm)--(vtm);\draw[line width=1.15pt](sigmat)--(vt);\draw[line width=1.15pt](sigmatp)--(vtp);
\end{tikzpicture}}}
\caption{(a) Example of directed acyclic graph. The node $x_3$ is a collider in the path $x_2,x_3,x_1$, but it is a non-collider in the path $x_2,x_4,x_3,x_1$. Adding a link from $x_4$ to $x_1$ would result in a cyclic graph. (b) Belief network representation of a hidden Markov model. Rectangular nodes indicate discrete variables, whilst oval nodes indicate discrete or continuous variables. Filled nodes indicate that the variables are observed.}
\label{fig:indep_HMM}
\end{center}
\end{figure}

\subsection*{Basic Graphical Model Definitions}
\emph{A \textbf{graphical model} is a graph in which
nodes represent random variables and (undirected or directed) links represent statistical dependencies among variables.\\[5pt]
A \textbf{path} from node $x_i$ to node $x_j$ is a sequence of linked nodes connecting
$x_i$ to $x_j$.\\[5pt]
A node $x_i$ with a direct link towards $x_j$ is called \textbf{parent}
of $x_j$. In this case, $x_j$ is called \textbf{child} of $x_i$.\\[5pt]
An undirected path from $x_i$ to $x_j$ has a
\textbf{collider} at $x_k$ if there are two arrows along the path pointing towards $x_k$.}
Notice that a node can be a collider in a path, but a non-collider in another path. For example, in \figref{fig:indep_HMM} (a) $x_3$ is a collider in the path $x_2,x_3,x_1$, but it is a non-collider in the path $x_2,x_4,x_3,x_1$.\\[5pt]
\emph{A \textbf{directed acyclic graph} is a graph with no
directed paths starting and ending at the same node.} For example, the directed graph in \figref{fig:indep_HMM} (a) is acyclic. Adding a link from $x_4$ to $x_1$ would make the graph cyclic.\\[5pt]
\emph{A \textbf{belief network} is a directed acyclic graph in which each node $x_i$ has associated
the conditional distribution $p(x_i|parents(x_i))$. The joint distribution of all nodes in
the graph is given by the product of all conditional distributions}
\begin{align*}
p(x_{1:D}\equiv x_1,\ldots,x_D)=\prod_{i=1}^Dp(x_i|parents(x_i)).
\end{align*}
\emph{A node $x_i$ is called \textbf{ancestor} of a node $x_j$ if there exists a directed path from $x_i$ to $x_j$. In this case, $x_j$ is called \textbf{descendant} of $x_i$.}

\subsection*{Assessing Statistical Independence}
\vskip0.1cm
\textbf{Method I.} Given the sets of random variables ${\cal X}, {\cal Y}$ and ${\cal Z}$,
${\cal X}$ and ${\cal Y}$ are statistically independent given ${\cal Z}$ (denoted with ${\cal X} \ci {\cal Y} \,|\, {\cal Z}$) if all paths
from any element of ${\cal X}$ to any element of ${\cal Y}$ are blocked. A \textbf{path is blocked} if at least one of the following conditions is satisfied
\begin{enumerate}
\item There is a non-collider in the path which is in the conditioning set ${\cal Z}$.
\item There is a collider in the path such that nor the collider nor any of its descendants is in ${\cal Z}$.
\end{enumerate}
\textbf{Method II.} This method consists of converting the directed graph into an undirected one and then using the rules of independence for undirected graphs. This is achieved by the following steps:
\begin{enumerate}
\item Create the ancestral graph: remove any node which is neither in ${\cal X}\cup{\cal Y}\cup {\cal Z}$ nor an ancestor of a node in this set, together with any links in or out of such nodes.
\item Perform moralisation: add a link between any two nodes which have a common child. Remove arrowheads.
\item Use independence rules for undirected graphs: if all paths which join a node in ${\cal X}$ to one in ${\cal Y}$
pass through any member of ${\cal Z}$ then ${\cal X}\ci{\cal Y} \,|\, {\cal Z}$.
\end{enumerate}

\section{Hidden Markov Model}
We start our analysis with the Hidden Markov Model (\HMM), which
represents the basic model from which extensions will be derived.
The HMM has a belief network representation given in \figref{fig:indep_HMM} (b), in which the discrete or continuous (possibly multivariate) variable $v_t$
and the discrete variable $s_t$ represent respectively the observations and the underlying dynamics regime at time $t$. The joint distribution
of all variables factorises as\footnote{We use $x_{1:T}$ as a shorthand for $x_1,\ldots,x_T$.}
\begin{align*}
p(v_{1:T},s_{1:T})=p(v_1|s_1)p(s_1)\prod_{t=2}^T p(v_t|s_t)p(s_t|s_{t-1}).
\end{align*}
We assume a homogeneous regime switching, parameterized by a vector $\tilde\pi$ and a matrix $\pi$ such that
$p(s_1)=\tilde\pi_{s_1}$ and $p(s_t|s_{t-1})=\pi_{s_ts_{t-1}}\textrm{ for }t>1$.
For the case in which $v_t$ is a continuous variable, the distribution $p(v_t|s_t)$ is commonly modelled
as a Gaussian mixture.

There are three main properties that can limit the modelling accuracy of \HMMs, namely
\begin{enumerate}
\item The strong independence assumption among observations $v_t \ci v_{{\setminus t}\equiv v_1,\ldots,v_{t-1},v_{t+1},\ldots,v_T} \,|\, s_t$,
with consequent lack of smoothness in modeling the observations.
\item The weak modelling of the state-duration distribution: the probability of observing
state $i$ for $d$ consecutive time-steps is implicitly given by the geometric distribution
$\pi_{ii}^{d-1}(1-\pi_{ii})$, which is often inappropriate and encouraging too fast regime switching. 
\item The limited power in modeling the observations when high noise is present.
In particular, the \HMM is unable to provide an estimate of the dynamics underlying noisy observations.
\end{enumerate}

In the next sections we describe extensions of the HMM that overcome these limitations.

\section{Extension of HMM to account for Dependency among Observations\label{sec:HMM_DO}}
\begin{figure}[t]
\renewcommand{\subfigcapskip}{0pt}
\begin{center}
\subfigure[]{\scalebox{0.65}{
\begin{tikzpicture}[dgraph]
\node[] at (-0.9,0) {$\cdots$};
\node[disc] (sigmatm2) at (0,0) {$s_{t-2}$};
\node[disc] (sigmatm) at (2,0) {$s_{t-1}$};
\node[disc] (sigmat) at (4,0) {$s_t$};
\node[disc] (sigmatp) at (6,0) {$s_{t+1}$};
\node[disc] (sigmatp2) at (8,0) {$s_{t+2}$};
\node[] at (8.9,0) {$\cdots$};
\node[ocont,obs] (vtm2) at (0,-1.5) {$v_{t-2}$};
\node[ocont,obs] (vtm) at (2,-1.5) {$v_{t-1}$};
\node[ocont,obs] (vt) at (4,-1.5) {$v_t$};
\node[ocont,obs] (vtp) at (6,-1.5) {$v_{t+1}$};
\node[ocont,obs] (vtp2) at (8,-1.5) {$v_{t+2}$};
\draw[line width=1.15pt](sigmatm2)--(sigmatm);\draw[line width=1.15pt](sigmatm2)--(vtm2);\draw[line width=1.15pt](vtm2)--(vtm);
\draw[line width=1.15pt](sigmatp)--(sigmatp2);\draw[line width=1.15pt](sigmatp2)--(vtp2);
\draw[line width=1.15pt](sigmatm)--(sigmat);\draw[line width=1.15pt](sigmat)--(sigmatp);
\draw[line width=1.15pt](sigmatm)--(vtm);\draw[line width=1.15pt](sigmat)--(vt);\draw[line width=1.15pt](sigmatp)--(vtp);
\draw[line width=1.15pt](vtm)--(vt);
\draw[line width=1.15pt](vt)--(vtp);
\draw[line width=1.15pt](vtp)--(vtp2);
\end{tikzpicture}}}
\subfigure[]{\scalebox{0.65}{
\begin{tikzpicture}[ugraph]
\node[] at (-0.9,0) {$\cdots$};
\node[disc] (sigmatm2) at (0,0) {$s_{t-2}$};
\node[disc] (sigmatm) at (2,0) {$s_{t-1}$};
\node[disc] (sigmat) at (4,0) {$s_t$};
\node[disc] (sigmatp) at (6,0) {$s_{t+1}$};
\node[disc] (sigmatp2) at (8,0) {$s_{t+2}$};
\node[] at (8.9,0) {$\cdots$};
\node[ocont,obs] (vtm2) at (0,-1.5) {$v_{t-2}$};
\node[ocont,obs] (vtm) at (2,-1.5) {$v_{t-1}$};
\node[ocont,obs] (vt) at (4,-1.5) {$v_t$};
\node[ocont,obs] (vtp) at (6,-1.5) {$v_{t+1}$};
\node[ocont,obs] (vtp2) at (8,-1.5) {$v_{t+2}$};
\draw[line width=1.15pt](sigmatm2)--(sigmatm);\draw[line width=1.15pt](sigmatm2)--(vtm2);\draw[line width=1.15pt](vtm2)--(vtm);
\draw[line width=1.15pt](sigmatp)--(sigmatp2);\draw[line width=1.15pt](sigmatp2)--(vtp2);
\draw[line width=1.15pt](sigmatm)--(sigmat);\draw[line width=1.15pt](sigmat)--(sigmatp);
\draw[line width=1.15pt](sigmatm)--(vtm);\draw[line width=1.15pt](sigmat)--(vt);\draw[line width=1.15pt](sigmatp)--(vtp);
\draw[line width=1.15pt](sigmatm)--(vtm2);
\draw[line width=1.15pt](sigmat)--(vtm);\draw[line width=1.15pt](sigmatp)--(vt);
\draw[line width=1.15pt](sigmatp2)--(vtp);
\draw[line width=1.15pt](vtm)--(vt);
\draw[line width=1.15pt](vt)--(vtp);
\draw[line width=1.15pt](vtp)--(vtp2);
\end{tikzpicture}}}
\caption{(a) Extension of the HMM to account for first-order Markovian dependence among the observations. (b) Ancestral moralized graph of the belief network in (a).}
\label{fig:HMMMark}
\end{center}
\end{figure}
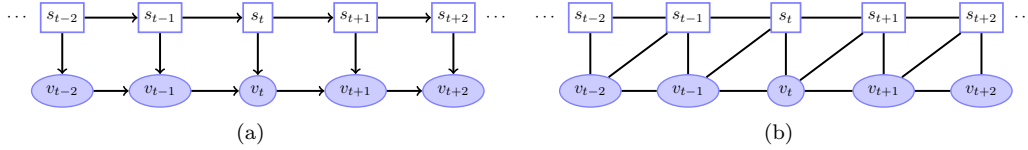

A simple way of relaxing the strong independence assumption $v_t \ci v_{\setminus t} \,|\, s_t$ in the \HMM~is to introduce Markovian dependence
among the observations. In the belief network, this is represented by adding links from past to current observations,
as shown in \figref{fig:HMMMark} (a) for the case of first-order dependence. For $k$-order dependence, the joint distribution can be
written as\footnote{We use the convention $x_{\tau}=\emptyset$ for ${\tau}\leq 0$.},\footnote{The \HMM~can be obtain as a special case when $k=0$, with the convention $x_{t':t''}=\emptyset$, for $t'>t''$.}
\begin{align*}
p(v_{1:T},s_{1:T})=\prod_{t=1}^T p(v_t|s_t,v_{t-k:t-1})p(s_t|s_{t-1}).
\end{align*}
Models of this type include, for example, the Switching Autoregressive Model (\SARM) in which\footnote{The distribution for $v_{1:k}$ has to be defined separately.} $v_t=f(s_t,v_{t-k:t-1})+\eta_t=\sum_{i=1}^k a^{s_t}_i v_{t-i}+\eta_t$
($\eta_t$ being a Gaussian or Gamma noise vector) \cite{hamilton89new,hamilton90analysis,
hamilton93estimation}, and extensions to a nonlinear interaction $f$ \cite{susmel00switching}.

In the sequel, we describe efficient routines for solving the two most common inference problems in these models, namely the computation of $p(s_t|v_{1:T})$, known as smoothing, and the estimation of the most likely sequence of states $s^*_{1:T}=\argmax p(s_{1:T}|v_{1:T})$.
\vskip-0.5cm
\subsection*{Smoothing with Parallel Routines}
The estimation of $\gamma^{s_t}_t\equiv p(s_t|v_{1:T})$ can be obtained as\footnote{The normalising constant $p(v_{1:T})$ is estimated as $\sum_{s_t} \alpha^{s_t}_T$.}\vspace{-0.2cm}\\
\begin{align*}
\gamma^{s_t}_t&\propto p(s_t,v_{1:T})
=\underbrace{p(v_{t+1:T}|s_t,\cancel{v_{1:t-k}},v_{t-k+1:t})}_{\beta^{s_t}_t}\underbrace{p(s_t,v_{1:t})}_{\alpha^{s_t}_t},
\end{align*}
where we have used the notation $p(v_{t+1:T}|s_t,\cancel{v_{1:t-k}},v_{t-k+1:t})$ to emphasise the independence relation $v_{t+1:T} \ci v_{1:t-k} \,|\, \{s_t,v_{t-k+1:t}\}$. This relation can be assessed using any of the two methods explained in
\secref{sec:GM}. For example, for $k=1$ we have $v_{t+1:T} \ci v_{1:t-1} \,|\, \{s_t,v_t\}$, which follows from
\begin{description}
\item[Method I:] All paths from (any element of) $v_{1:t-1}$ to
(any element of) $v_{t+1:T}$ are blocked, as $v_{t+1:T}$ are reached by passing (1) from both $s_t$ and $v_t$, (2) from
$s_t$ only, (3) from $v_t$ only (\figref{fig:HMMMark}(a)).
In cases (1) and (2), $s_t$ is a non-collider in the path which is in the conditioning set, and therefore condition 1. is satisfied.
In case (3), $v_t$ is a non-collider in the path which is in the conditioning set, and therefore condition 1. is satisfied.\\
\item[Method II:] The ancestral moralised graph is depicted in \figref{fig:HMMMark}(b). All paths from $v_{1:t-1}$ to $v_{t+1:T}$ have to pass through
an element of the conditioning set $s_t$ or $v_t$.
\end{description}
The terms $\alpha^{s_t}_t$ and $\beta^{s_t}_t$ can be recursively computed as follows\footnote{The initialisation is given by $\alpha^{s_1}_1=p(v_1|s_1)\tilde\pi_{s_1}$ and $\beta^{s_T}_T=1$.}\vspace{-0.2cm}\\
\begin{align*}
\alpha^{s_t}_t&=p(v_t|s_t,\cancel{v_{1:t-k-1}},v_{t-k:t-1})p(s_t,v_{1:t-1})\\[4pt]
&=p(v_t|s_t,v_{t-k:t-1})\sum_{s_{t-1}}p(s_t|s_{t-1},\cancel{v_{1:t-1}})p(s_{t-1},v_{1:t-1})\\[0pt]
&=p(v_t|s_t,v_{t-k:t-1})\sum_{s_{t-1}}\pi_{s_ts_{t-1}}\alpha^{s_{t-1}}_{t-1},\\[7pt]
\beta^{s_t}_t&=\sum_{s_{t+1}}p(v_{t+1:T}|\cancel{s_t},s_{t+1},v_{t-k+1:t})p(s_{t+1}|s_t,\cancel{v_{t-k+1:t}})\\
&=\sum_{s_{t+1}}p(v_{t+2:T}|s_{t+1},v_{\cancel{t-k+1},{t-k+2:t+1}})p(v_{t+1}|s_{t+1},v_{t-k+1:t})\pi_{s_{t+1}s_t}\\
&=\sum_{s_{t+1}}\beta^{s_{t+1}}_{t+1}p(v_{t+1}|s_{t+1},v_{t-k+1:t})\pi_{s_{t+1}s_t}.
\end{align*}
For $k=1$ for example, the independence relation $s_t \ci v_{1:t-1} \,|\, s_{t-1}$
can be demonstrated by noticing that all paths from $v_{1:t-1}$ to $s_t$ reach $s_t$ from (1) the non-collider
$s_{t-1}$ which is in the conditioning set, (2) the collider $v_t$ which (together with all its
descendants) is not in the conditioning set, (3) $s_{t+1}$ which imposes passing through
a collider (e.g. $v_{t+1}$) which (together with all its descendants) is not in the conditioning set.

Notice that the $\alpha^{s_t}_t$ and $\beta^{s_t}_t$ recursions can be performed in parallel, after which $\gamma^{s_t}_t$ is computed.
The computational cost of these recursions is $O(TS^2)$ (without considering the cost of estimating terms such as $p(v_t|s_t,v_{t-k:t-1})$).
In order to avoid numerical overflow/underflow problems, computations are often preformed in log scale.

\paragraph{Smoothing with Sequential Routines}
An alternative way of performing smoothing is to first compute the filtered distribution $\alpha^{s_t}_t\equiv p(s_t|v_{1:t})$ and then use
this estimate to compute $\gamma^{s_t}_t\equiv p(s_t|v_{1:T})$ as follows\\
\scalebox{0.9}{
\begin{minipage}{0.45\textwidth}
\begin{align*}
\alpha^{s_t}_t&\propto p(v_t|s_t,\cancel{v_{1:t-k-1}},v_{t-k:t-1})p(s_t|v_{1:t-1})\\[3pt]
&=p(v_t|s_t,v_{t-k:t-1})\sum_{s_{t-1}}p(s_t|s_{t-1},\cancel{v_{1:t-1}})p(s_{t-1}|v_{1:t-1})\\
&=p(v_t|s_t,v_{t-k:t-1})\sum_{s_{t-1}}\pi_{s_ts_{t-1}}\alpha^{s_{t-1}}_{t-1},
\end{align*}
\end{minipage}
\begin{minipage}{0.45\textwidth}
\begin{align*}
\gamma^{s_t}_t&=\sum_{s_{t+1}}p(s_t|s_{t+1},v_{1:t},\cancel{v_{t+1:T}})p(s_{t+1}|v_{1:T})\\
&=\sum_{s_{t+1}}\frac{p(s_{t+1}|s_t,\cancel{v_{1:t}})p(s_t|v_{1:t})}{\sum_{\tilde s_t}p(s_{t+1}|\tilde s_t,\cancel{v_{1:t}})p(\tilde s_t|v_{1:t})}\gamma^{s_{t+1}}_{t+1}\\
&=\sum_{s_{t+1}}\frac{\pi_{s_{t+1}s_t}\alpha^{s_t}_t}{\sum_{\tilde s_t}\pi_{s_{t+1}\tilde s_t}\alpha^{\tilde s_t}_t}\gamma^{s_{t+1}}_{t+1},
\end{align*}
\end{minipage}}
\vskip0.1cm
where the normalisation $p(v_t|v_{1:t-1})$ in the $\alpha^{s_t}_t$ recursion is obtained by summing the rhs of the above over all values of $s_t$.
Notice that we are not making use of the observations after the $\alpha^{s_t}_t$ have been computed. These routines do not require working in a log scale.

In the following sections, we will focus on the parallel smoothing approach and come back a sequential smoothing approach in \secref{sec:SLGSSMM1M2}.

\subsection*{Computing the Most Likely Sequence of Hidden Variables}
If we define $\delta^{s_t}_t=\max_{s_{1:t-1}} p(s_{1:t},v_{1:t})$, the most likely sequence of hidden variables $s^*_{1:T}=\argmax_{s_{1:T}} p(s_{1:T}|v_{1:T})$ can be obtained with the following algorithm
\begin{align*}
&\delta^{s_1}_1=p(s_1,v_1)=\alpha^{s_1}_1\\
&\textrm{for } t=2,\ldots,T \hspace{0.2cm} \delta^{s_t}_t=p(v_t|s_t,v_{t-k:t-1})\max_{s_{t-1}} \pi_{s_ts_{t-1}}\delta^{s_{t-1}}_{t-1},\hspace{0.3cm}\psi^{s_t}_t=\argmax_{s_{t-1}} \pi_{s_ts_{t-1}}\delta^{s_{t-1}}_{t-1}\\[-1pt]
&s^*_T=\argmax_{s_T} \delta^{s_T}_T, \hspace{0.3cm}\textrm{for } t=T-1,\ldots,1 \hspace{0.2cm} s^*_t=\psi^{s^*_{t+1}}_{t+1}
\end{align*}
where the recursion for $\delta^{s_t}_t$ is obtained as the $\alpha^{s_t}_t$ recursion with the sum replaced by the max operator.

\subsection*{Artificial Data Example}
In this section, we illustrate how the switching autoregressive model (\SARM) can overcomes the limitation of the \HMM~in
modeling the temporal structure of a continuous time-series.
In this continuous case, the observations in the \HMM~are commonly modeled as mixture of $M$ Gaussians, for which the smoothed distributions
are given by $\gamma^{s_t,m_t}_t=p(v_{t+1:T}|s_t,\cancel{m_t},\cancel{v_{1:t}})p(s_t,m_t,v_{1:t})$ with
\begin{align*}
\alpha^{s_t,m_t}_t&=p(s_t,m_t,v_{1:t})\\
&=p(v_t|s_t,m_t,\cancel{v_{1:t-1}})p(m_t|s_t,\cancel{v_{1:t-1}})\sum_{s_{t-1}}p(s_t|s_{t-1},\cancel{v_{1:t-1}})p(s_{t-1},v_{1:t-1})\\
&=p(v_t|s_t,m_t)p(m_t|s_t)\sum_{s_{t-1}}p(s_t|s_{t-1})\sum_{m_{t-1}}p(s_{t-1},m_{t-1}|v_{1:t-1}),
\end{align*}
which has computational cost $O(TS(M+SM))$, and
\begin{align*}
\beta^{s_t}_t&=p(v_{t+1:T}|s_t)\\
&=\sum_{s_{t+1}}p(v_{t+2:T}|\cancel{s_t},s_{t+1},\cancel{v_{t+1}})\sum_{m_{t+1}}p(v_{t+1}|\cancel{s_t},s_{t+1},m_{t+1})p(m_{t+1}|\cancel{s_t},s_{t+1})p(s_{t+1}|s_t)
\end{align*}
which has computational cost $O(TS^2M)$.

\begin{figure}[t]
\begin{center}
\hspace{-0.2cm}
\includegraphics[]{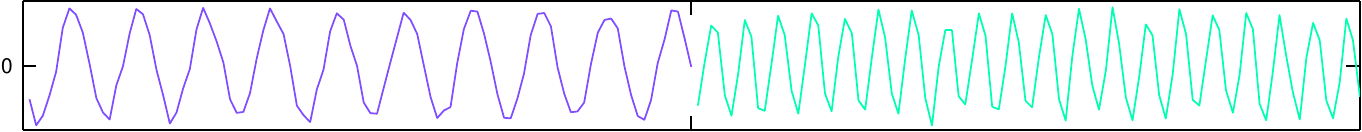}\\
\includegraphics[]{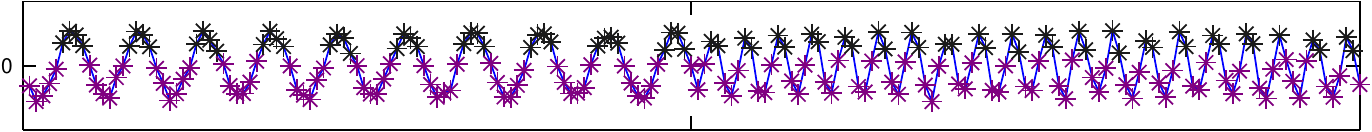}\\
\includegraphics[]{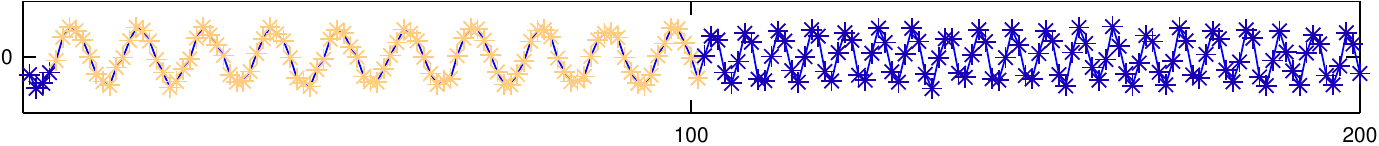}
\end{center}
\caption{Top panel: the generated time-series up to the first 20 regime switches. Bottom panel: the sequence of (red) underlying and (gray) estimated regimes as $\argmax_{s_t}\sum_{c_t}\gamma^{s_t,c_t}_t$. The intensity of the gray indicates the value $\max_{s_t}\sum_{c_t}\gamma^{s_t,c_t}_t$ (darker colour means greater value).}
\label{fig:example_hmm_sar}
\end{figure}

We generated a 200 time-step-long time-series, corresponding to a noisy sinusoid with switching frequency at $t=100$. The time-series with the correct segmentation represented by different colours is plotted on the top of \figref{fig:example_hmm_sar}.

We then learned the parameters of a \HMM~with observations modeled as mixtures of $M=3$ Gaussians, using an EM algorithm where the M-step updates are given by
\begin{align*}
&\mu_{s_t,m_t}=\frac{\sum_{t=1}^Tv_tp(s_t,m_t|v_{1:T},\Theta^{i-1})}{\sum_{t=1}^Tp(s_t,m_t|v_{1:T},\Theta^{i-1})}\\
&\Sigma_{s_t,m_t}=\frac{\sum_{t=1}^T(v_t-\mu_{s_t,m_t})(v_t-\mu_{s_t,m_t})\trans p(s_t,m_t|v_{1:T},\Theta^{i-1})}{\sum_{t=1}^Tp(s_t,m_t|v_{1:T})}\\
&p(s_1)=p(s_1|v_{1:T},\Theta^{i-1}), p(s_t|s_{t-1})=\frac{\sum_{t=2}^Tp(s_{t-1:t}|v_{1:T},\Theta^{i-1})}{\sum_{t=2}^T\sum_{s_t}p(s_{t-1:t}|v_{1:T},\Theta^{i-1})}\\
&p(m_t|s_t)=\frac{\sum_{t=1}^Tp(s_t,m_t|v_{1:T},\Theta^{i-1})}{\sum_{t=1}^T\sum_{m_t}p(s_t,m_t|v_{1:T},\Theta^{i-1})}.
\end{align*}
The resulting segmentation is given in the middle of \figref{fig:example_hmm_sar}. As we can see,
the \HMM~splits the time-series into two regimes, one corresponding to higher values and the other one corresponding to lower values
of the time-series. It is clear from this example, that in order to obtain the desired segmentation more temporal structure needs to be encoded into the model. One solution would be to add a link from $m_{t-1}$ to $m_t$, which would give an $\alpha$ routine
\begin{align*}
\alpha^{s_t,m_t}_t&=p(s_t,m_t,v_{1:t})\\
&=p(v_t|s_t,m_t,\cancel{v_{1:t-1}})\sum_{m_{t-1}}p(m_{t-1:t},s_t,v_{1:t-1})\\
&=p(v_t|s_t,m_t)\sum_{m_{t-1}}p(m_t|m_{t-1},s_t,\cancel{v_{1:t-1}})\sum_{s_{t-1}}p(m_{t-1},s_{t-1:t},v_{1:t-1})\\
&=p(v_t|s_t,m_t)\sum_{m_{t-1}}p(m_t|m_{t-1},s_t)\sum_{s_{t-1}}p(s_t|s_{t-1},\cancel{m_{t-1}},\cancel{v_{1:t-1}})p(s_{t-1},m_{t-1},v_{1:t-1})\\
&=p(v_t|s_t,m_t)\sum_{m_{t-1}}p(m_t|m_{t-1},s_t)\sum_{s_{t-1}}p(s_t|s_{t-1})\alpha^{s_{t-1},m_{t-1}}_{t-1},
\end{align*}
which has computational cost $O(TSM(S+M))$. This therefore would increase computational complexity considerably for complex time-series in which $M$
needs to be high. A less expensive alternative is to use a \SARM, which. After learning the parameters similarly to the \HMM~case, \SARM gives the
desired segmentation as shown in the bottom of \figref{fig:example_hmm_sar}.


\section{Extension of the HMM to Model Explicitly the Regime-Duration Distribution\label{sec:M1M2}}
A way of overcoming the limitation of the implicit geometric regime-duration distribution of \HMMs, is to define a different
distribution by introducing extra random variables, at the price of increasing the computational cost of inference.
This idea was first presented in \cite{ferguson80variable} and largely followed in the speech \cite{murphy2002hsm,ostendorf96from,rabiner89tutorial} and statistics communities \cite{guedon03estimating,samsom01fitting}. There are two main ways to achieved this, which differ by
using one or two sets of discrete hidden variables respectively. In the sequel we introduce them and discuss their relative properties.

In the exposition, we will pay particular attention to the calculation of the effective computational cost of the inference routines,
with the goal of clarifying incorrectness and misunderstanding of the current literature.

\subsection{Modelling the Regime-Duration Distribution with One Set of Duration-Count Variables \label{sec:M1}}
When using one set of discrete hidden variables $c_{1:T}$ for defining an explicit regime-duration distribution, we can think of two different approaches for
modeling such variables. In the first approach, $c_t$ provides information about the time-step $\tau\geq t$ at which the current regime ends,
which gives rise to decreasing duration-count variables within a regime. In the second approach, $c_t$ provides information about the time-step $\tau\leq t$ at which the current starts, which gives rise to increasing duration-count variables within a regime. These two approaches and their properties are discussed in detail in the sequel.

\subsubsection{Decreasing Duration-Count Variables within a Regime \label{sec:M1st}}
In the first approach to explicit modeling of the regime-duration distribution using decreasing duration-count variables within a regime, at a beginning of a regime $t$, $c_t$ indicates the number of time-steps (duration) spanned by the regime sampled according to a regime-duration distribution. At subsequent time-steps, the duration-count variables take progressively smaller values, until reaching value 1 at the end of the regime. More specifically, using the notation $\sigma_t=\{s_t,c_t\}$, we define $p(\sigma_{1:T})=\prod_t p(\sigma_t|\sigma_{t-1})=\prod_t p(s_t|s_{t-1},c_{t-1})p(c_t|c_{t-1})$
with

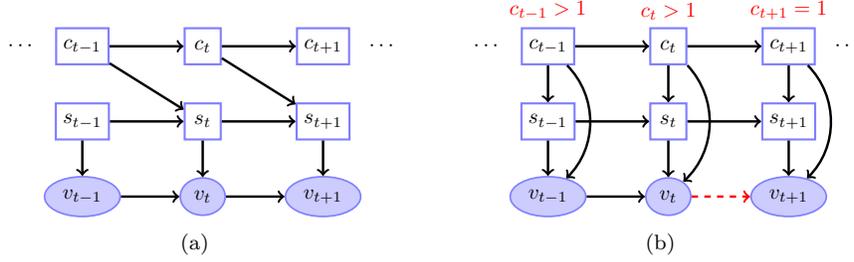
\begin{figure}[t]
\renewcommand{\subfigcapskip}{0pt}
\begin{center}
\subfigure[]{\scalebox{0.8}{
\begin{tikzpicture}[dgraph]
\node[] at (1,0) {$\cdots$};
\node[disc] (sigmatm) at (2,0) {$c_{t-1}$};
\node[disc] (sigmat) at (4,0) {$c_t$};
\node[disc] (sigmatp) at (6,0) {$c_{t+1}$};
\node[] at (7,0) {$\cdots$};
\node[disc] (stm) at (2,-1.25) {$s_{t-1}$};
\node[disc] (st) at (4,-1.25) {$s_t$};
\node[disc] (stp) at (6,-1.25) {$s_{t+1}$};
\node[ocont,obs] (vtm) at (2,-2.5) {$v_{t-1}$};
\node[ocont,obs] (vt) at (4,-2.5) {$v_t$};
\node[ocont,obs] (vtp) at (6,-2.5) {$v_{t+1}$};
\draw[line width=1.15pt](sigmatm)--(sigmat);\draw[line width=1.15pt](sigmat)--(sigmatp);
\draw[line width=1.15pt](stm)--(st);\draw[line width=1.15pt](st)--(stp);
\draw[line width=1.15pt](sigmatm)--(st);\draw[line width=1.15pt](sigmat)--(stp);
\draw[line width=1.15pt](stm)--(vtm);\draw[line width=1.15pt](st)--(vt);\draw[line width=1.15pt](stp)--(vtp);
\draw[line width=1.15pt](vtm)--(vt);
\draw[line width=1.15pt](vt)--(vtp);
\end{tikzpicture}}}
\hspace{0.5cm}
\subfigure[]{\scalebox{0.8}{
\begin{tikzpicture}
\node[] at (1,0) {$\cdots$};
\node[disc] (sigmatm) at (2,0) {$c_{t-1}$};
\node[disc] (sigmat) at (4,0) {$c_t$};
\node[disc] (sigmatp) at (6,0) {$c_{t+1}$};
\node[disc] (stm) at (2,-1.25) {$s_{t-1}$};
\node[disc] (st) at (4,-1.25) {$s_t$};
\node[disc] (stp) at (6,-1.25) {$s_{t+1}$};
\node[] at (7,0) {$\cdots$};
\node[ocont,obs] (vtm) at (2,-2.5) {$v_{t-1}$};
\node[ocont,obs] (vt) at (4,-2.5) {$v_t$};
\node[ocont,obs] (vtp) at (6,-2.5) {$v_{t+1}$};
\node [red,above] at (sigmatm.north) {$c_{t-1}>1$};
\node [red,above] at (sigmat.north) {$c_t>1$};
\node [red,above] at (sigmatp.north) {$c_{t+1}=1$};
\draw[->, line width=1.15pt](stm)--(st);\draw[->, line width=1.15pt](st)--(stp);
\draw[->, line width=1.15pt](stm)--(vtm);\draw[->, line width=1.15pt](st)--(vt);\draw[->, line width=1.15pt](stp)--(vtp);
\draw[->, line width=1.15pt](sigmatm)--(stm);\draw[->, line width=1.15pt](sigmat)--(st);\draw[->, line width=1.15pt](sigmatp)--(stp);
\draw[->, line width=1.15pt](sigmatm)--(sigmat);\draw[->, line width=1.15pt](sigmat)--(sigmatp);
\draw[->, line width=1.15pt](sigmatm)to [bend left=45](vtm);\draw[->, line width=1.15pt](sigmat)to [bend left=45](vt);\draw[->, line width=1.15pt](sigmatp)to [bend left=45](vtp);
\draw[->,line width=1.15pt](vtm)--(vt);\draw[->,red,dashed,line width=1.15pt](vt)--(vtp);
\end{tikzpicture}}}
\end{center}
\caption{\HMSMs~in which the regime-duration distribution is explicitly modelled using a set of discrete variables $c_{1:T}$ (a) as in \secref{sec:M1st} and (b) as in \secref{sec:M1alt}.
In (b) the visible dependence across regimes (from time $t$ to time $t+1$) is cut (red dashed lines) ($c_{t+1}=1$ indicates the start of a new regime at time $t+1$). 
}
\label{fig:M1}
\end{figure}

\begin{minipage}{0.45\textwidth}
\begin{align*}
p(s_t|s_{t-1},c_{t-1})&=\begin{cases}
	\pi_{s_ts_{t-1}} 	& \textrm{if } c_{t-1}\!=\!1,\\
	\delta_{s_t=s_{t-1}} \hspace{0.27cm}	& \textrm{if } c_{t-1}\!>\!1,
\end{cases}
\end{align*}
\end{minipage}
\hspace{0.5cm}
\begin{minipage}{0.45\textwidth}
\begin{align*}
p(c_t|c_{t-1})&=\begin{cases}
	\rho_{c_t} 		& \textrm{if } c_{t-1}\!=\!1,\\
	\delta_{c_t=c_{t-1}-1} 	& \textrm{if } c_{t-1}\!>\!1,
\end{cases}
\end{align*}
\end{minipage}

where $\delta$ is the Dirac delta and $\rho$ specifies the state-duration distribution. By considering a distribution that is zero outside the interval $\{d_{\min},\ldots,d_{\max}\}$, we can impose constraints on the minimum and (if $\pi_{ii}=0$) maximum duration allowed for a regime. We consider the case of $k$-order Markovian dependence among the observations, and therefore obtain a belief network representation of the model as in \figref{fig:M1} (a) (for the case $k=1$).
\subsection*{Smoothing with parallel Routines} To estimate $\gamma^{\sigma_t}_t=p(\sigma_t|v_{1:T})\propto p(v_{t+1:T}|\sigma_t,v_{t-k+1:t})p(\sigma_t,v_{1:t})=\beta^{\sigma_t}_t\alpha^{\sigma_t}_t$ we can use a similar approach to the one described in \secref{sec:HMM_DO}. Specifically\footnote{The initialisation is given by $\alpha^{\sigma_1}_1=p(v_1|s_1)\tilde\pi_{s_1}p(\textrm{duration}\geq c_1)=p(v_1|s_1)\tilde\pi_{s_1}\sum_{i=c_1}^{d_{\max}}\rho_i$. Notice that this way we are not imposing that the first regime starts at the first time-step $t=1$.}\vspace{-0.2cm}\\
\begin{align*}
\alpha^{\sigma_t}_t&=p(v_t|s_t,\cancel{c_t},\cancel{v_{1:t-k-1}},v_{t-k:t-1})p(\sigma_{t},v_{1:t-1})\\[3pt]
&=p(v_t|s_t,v_{t-k:t-1})\sum_{\sigma_{t-1}}p(\sigma_t|\sigma_{t-1},\cancel{v_{1:t-1}})\alpha^{\sigma_{t-1}}_{t-1}\\[-3pt]
&=p(v_t|s_t,v_{t-k:t-1})\Big\{\delta_{c_t<d_{\max}}\alpha^{s_t,c_t+1}_{t-1}+\rho_{c_t}\sum_{s_{t-1}}\pi_{s_ts_{t-1}}\alpha^{s_{t-1},1}_{t-1}\Big\},
\end{align*}
where, by exploiting the fact that $c_t<d_{\max}$ implies that either $c_{t-1}=c_t+1, s_{t-1}=s_t$ or $c_{t-1}=1$, and the fact that $c_t=d_{\max}$ implies that $c_{t-1}=1$, we reduced the cost\footnote{We do not consider the cost of computing $p(v_t|s_t,v_{t-k:t-1})$.} from
$O(TS^2d^2_{\max})$ to $O(TS^2d_{\max})$. Further saving may be obtained by pre-computing $\sum_{s_{t-1}}\pi_{s_ts_{t-1}}\alpha^{s_{t-1},1}_{t-1}$, giving a complexity of
$O(TS(S+d_{\max}))$. This cost is however more expensive than the cost $O(TS^2)$ of the model in \secref{sec:HMM_DO}.

The term $\beta^{\sigma_t}_t$ can be obtained as follows\footnote{The initialisation is given by $\beta^{s_T,c_T}_T=1$.}
\begin{align*}
\!\beta^{\sigma_t}_t&\!=\!\sum_{\sigma_{t+1}}p(v_{t+1:T}|\cancel{\sigma_t},\sigma_{t+1},v_{t-k+1:t})p(\sigma_{t+1}|\sigma_t,\cancel{v_{t-k+1:t}})\\
\!&\!=\!\sum_{\sigma_{t+1}}p(v_{t+2:T}|\sigma_{t+1},\cancel{v_{t-k+1}},v_{t-k+2:t+1})p(v_{t+1}|s_{t+1},\cancel{c_{t+1}},v_{t-k+1:t})p(\sigma_{t+1}|\sigma_t)\\[-7pt]
\!&\!=\!\delta_{c_t=1}\!\!\sum_{s_{t+1}}\!p(v_{t+1}|s_{t+1},v_{t-k+1:t})\pi_{s_{t+1}s_t}\hspace{-0.43cm}\sum_{c_{t+1}=d_{\min}}^{d_{\max}}\hspace{-0.4cm}\beta^{\sigma_{t+1}}_{t+1}\!\rho_{c_{t+1}}
\!+\!\delta_{c_t>1}p(v_{t+1}|s_{t+1},v_{t-k+1:t})\beta^{s_t,c_t-1}_{t+1},
\end{align*}
where, by using the fact that $c_t=1$ implies change of regime at time $t+1$, whilst $c_t>1$ implies $c_{t+1}=c_t-1$, $s_{t+1}=s_t$, we have reduced the cost to $O(TSd_{\max})$.

\subsection*{Most Likely Sequence of Hidden Variables} With the notation $\delta^{\sigma_t}_t=\max_{\sigma_{1:t-1}} p(\sigma_{1:t},v_{1:t})$, the most likely sequence $\sigma^*_{1:T}=\argmax_{\sigma_{1:T}} p(\sigma_{1:T}|v_{1:T})$ can be obtained as\\
\begin{align*}
&\delta^{\sigma_1}_1=p(\sigma_1,v_1)=\alpha^{\sigma_1}_1\\[3pt]
&\textrm{for }t=2,\cdots,T\\[1pt]
&\delta^{\sigma_t}_t=
\begin{cases}
p(v_t|s_t,v_{t-k:t-1})\max [\delta^{s_t,c_t+1}_{t-1},\rho_{c_t}\max_{s_{t-1}}\pi_{s_ts_{t-1}}\delta^{s_{t-1},1}_{t-1}]\hspace{0.07cm}	 & \textrm{if } c_t<d_{\max}\\
p(v_t|s_t,v_{t-k:t-1})\rho_{c_t}\max_{s_{t-1}}\pi_{s_ts_{t-1}}\delta^{s_{t-1},1}_{t-1} & \textrm{otherwise }
\end{cases}\\
&\psi^{\sigma_t}_t=\begin{cases}
	\{\argmax_{s_{t-1}}\pi_{s_t,s_{t-1}}\delta^{s_{t-1},1}_{t-1},1\} \hspace{0.07cm}	& \textrm{if } \rho_{c_t}\max_{s_{t-1}}\pi_{s_ts_{t-1}}\delta^{s_{t-1},1}_{t-1} > \delta^{s_t,c_t+1}_{t-1}\textrm{or } c_t=d_{\max}\\
\{s_t,c_t+1\} & \textrm{otherwise }
\end{cases}\\
&\sigma^*_T=\argmax_{\sigma_T} \delta^{\sigma_T}_T, \hspace{0.3cm}\textrm{for } t=T-1,\ldots,1 \hspace{0.2cm} \sigma^*_t=\psi^{\sigma^*_{t+1}}_{t+1}.
\end{align*}

\subsection*{Artificial Data Example}

\begin{figure}[t]
\begin{center}
\includegraphics[]{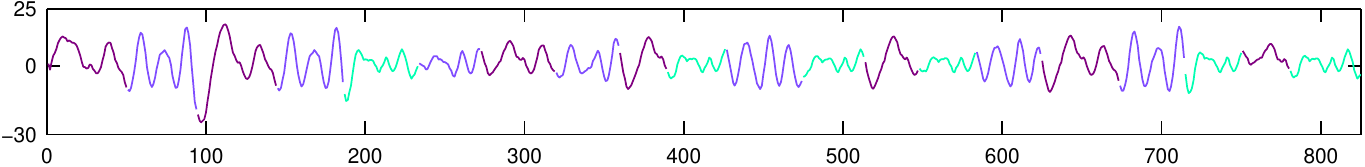}\vspace{-0.1cm}\\
\end{center}
\caption{The generated time-series up to the first 20 regime switches, indicated by different colours.}
\label{fig:sar_sardur_obs}
\end{figure}

There is considerable evidence in the literature that extensions of the \HMM~to explicitly model the regime-duration distribution give improvement in performance in several real-world problems such as handwriting and activity recognition, speech, MRI, financial time-series, rainfall time-series, and DNA analysis. Some specific examples can be found in \cite{yu10hidden}. In this section, we give a simple illustration of this property for the switching autoregressive model (\SARM), using artificial data.

We generated a time-series from a 3-regime, 3-order \SARM~$v_t=\sum_{i=1}^3 a^{s_t}_i v_{t-i}+\eta_t$ with\vspace{-0.2cm}\\
\begin{align*}
&a^1=(1.8, -0.99, 0),\hspace{0.15cm}a^2=(1.65, -0.9, 0.1),\hspace{0.15cm}a^3=(1.8, -0.85, 0);\hspace{0.35cm}\eta_t\sim{\cal N}(0,1).\\[-3pt]
\end{align*}
The time-series contains 100 regime switches of type and duration uniformly
sampled between 1 and 3 and between $d_{\min}=30$ and $d_{\max}=50$ respectively,
giving a total length of $T=4004$. Notice that this problem is quite hard, since the autoregressive coefficients for the three regimes are similar.
In \figref{fig:sar_sardur_obs} we show the generated time-series up to the first 20 regime switches, indicated by different colours.

\begin{figure}
\begin{center}
\vspace{-0.5cm}
\includegraphics[]{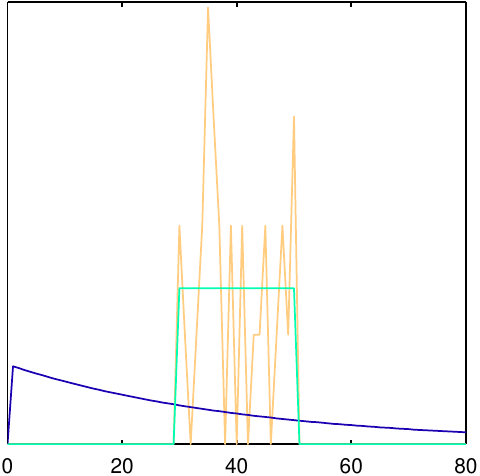}
\end{center}
\caption{The empirical (orange), the \GSARM~(blue), and \USARM~ (green) regime-duration distributions for regime 2.}
\label{fig:sar_sardur_dist}
\end{figure}

As a measure of segmentation error, we used the discrepancy between the correct and the estimated sequence of regimes (obtained as $\argmax_{s_t}\sum_{c_t}\gamma^{s_t,c_t}_t$ for the case of smoothing). For the \SARM~with geometric regime-duration distribution (\GSARM), we used the maximum likelihood value of the regime-switching distribution $\pi$ on the same sequence computed assuming that the other parameters and the segmentation are known.
For the \SARM~with explicit regime-duration distribution (\USARM), we used a uniform regime-duration distribution in the interval $\{30,\ldots,50\}$, and
$\pi_{ii}=0, \pi_{ji}=1/(S-1), j\neq i$.
In \figref{fig:sar_sardur_dist} we plot, for regime 2, the empirical regime-duration distribution of the time-series (orange) and the geometric
(blue) and uniform (green) distributions used in the \GSARM~and \USARM.

\begin{figure}[t]
\subfigure{
\hspace{-0.2cm}
\includegraphics[]{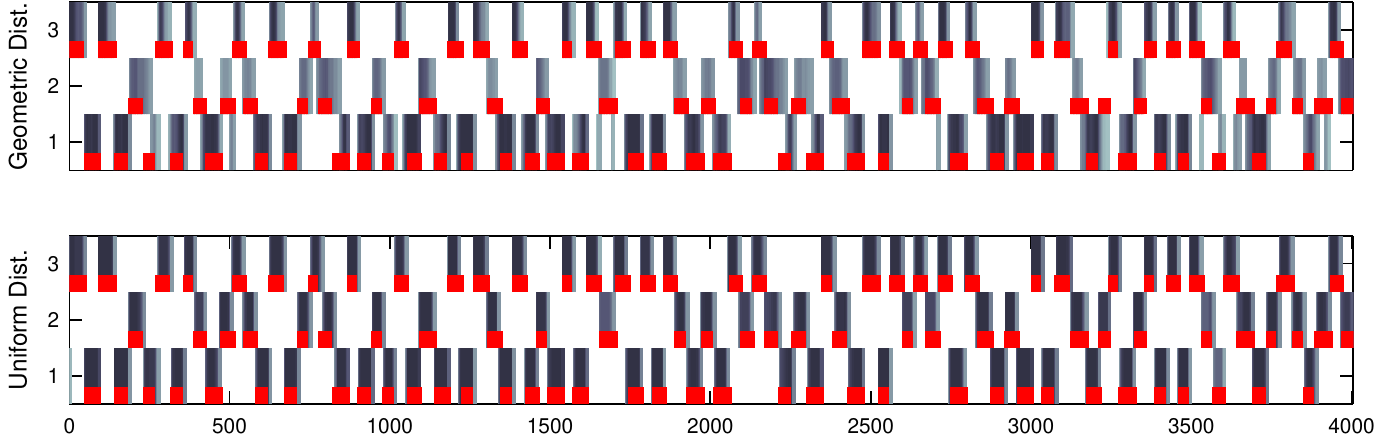}}
\caption{Top panel: the generated time-series up to the first 20 regime switches. Bottom panel: the sequence of (red) underlying and (gray) estimated regimes as $\argmax_{s_t}\sum_{c_t}\gamma^{s_t,c_t}_t$. The intensity of the gray indicates the value $\max_{s_t}\sum_{c_t}\gamma^{s_t,c_t}_t$ (darker colour means greater value).}
\label{fig:sar_sardur_gamma}
\end{figure}

We have performed two experiments.
In the first, the autoregressive coefficients and the Gaussian noise variance were assumed to be known, and therefore we measure difference in performance due to inference only.
In this case, the \GSARM~gave a segmentation error of $0.16\%$ for the case in which smoothing was employed and of $0.31\%$ for the case in which the most likely sequence of regimes was estimated, whilst the \USARM~gave a smaller error of $0.07\%$ and $0.08\%$. 
In the bottom panel of \figref{fig:sar_sardur_gamma} we show the sequence of (red) underlying and (gray) estimated regimes as $\argmax_{s_t}\sum_{c_t}\gamma^{s_t,c_t}_t$. The intensity of the gray indicates the value $\max_{s_t}\sum_{c_t}\gamma^{s_t,c_t}_t$ (darker colour indicates greater value). From this figure, we can observe that the \GSARM~is in general more uncertain about the regimes than the \USARM, particularly for regime 2 to which portions of time-series belonging to other regimes are often assigned.

In the second experiment, the autoregressive coefficients and the Gaussian noise variance were assumed to be unknown, giving rise to a more difficult task. These parameters were estimated
by the EM algorithm, using as initialisation
\begin{align*}
&a^1=(0.8, -0.99, 0),\hspace{0.15cm}a^2=(-0.65, 0.2, 0.1),\hspace{0.15cm}a^3=(0.9, -0.35, -0.3);\hspace{0.35cm}\eta_t\sim{\cal N}(0,100).\\[-3pt]
\end{align*}
In this case, the \GSARM~gave a segmentation error of $0.15\%$ for the case in which smoothing was employed and of $0.28\%$ for the case in which the most likely sequence of regimes was estimated, whilst the \USARM~gave a smaller error of $0.07\%$ and $0.07\%$.


\subsubsection{Increasing Duration-Count Variables within a Regime \label{sec:M1alt}}
An alternative model for the duration-count variables $c_{1:T}$ to the one described in \secref{sec:M1st}, would be to set $c_t=1$ at a beginning of a regime $t$, and define progressively increasing duration-count variables at subsequent time-steps within the regime. Specifically

\begin{minipage}{0.45\textwidth}
\begin{align*}
p(s_t|s_{t-1},c_t)&=\begin{cases}
	\pi_{s_ts_{t-1}} 	& \textrm{if } c_t\!=\!1,\\
	\delta_{s_t=s_{t-1}} \hspace{0.27cm}	& \textrm{if } c_t\!>\!1,
\end{cases}
\end{align*}
\end{minipage}
\hspace{0.5cm}
\begin{minipage}{0.45\textwidth}
\begin{align*}
p(c_t|c_{t-1})&=\begin{cases}
	\lambda_{c_{t-1}} & \textrm{if } c_t\!=\!c_{t-1}+1,\\	
	1-\lambda_{c_{t-1}} \hspace{0.27cm}	& \textrm{if } c_t=1,\\
    0 & \textrm{otherwise},
\end{cases}
\end{align*}
\end{minipage}

The relation between this model for $c_{1:T}$ and the one in \secref{sec:M1st} is given by
\begin{align*}
1-\lambda_{c_t}=\rho_{c_t}/\sum_{i=c_t}^{d_{\max}}\rho_i\,.
\end{align*}
This can be seen as follows
\begin{align*}
\rho_d&=p(\textrm{duration}= d)=(1-\lambda_d)\underbrace{\lambda_{d-1},\ldots,\lambda_2\lambda_1}_{p(\textrm{duration}\geq d)}=(1-\lambda_d)\sum_{i=d}^{d_{\max}}\rho_i\,.
\end{align*}

This alternative model for $c_{1:T}$ is useful to derive efficient inference routines, for example, for the case of Markovian dependence among observations in which we want to define independence across regimes (these types of models are often called change-point models). This can be achieved by adding a link from $c_t$ to $v_t$ in the belief network
(see \figref{fig:M1} (b) for the case of order $k=1$). Indeed, in this case the variable $c_t$ encodes information about the time-step at which the current regime started and therefore about when to cut past dependence, i.e. $p(v_t|\cancel{v_{t-k:t-c_t}},v_{t-c_t+1:t-1},s_t,c_t)$, $k\geq c_t$. This would not be possible with the model of \secref{sec:M1st}, in which at time $t$ information about when the current regime started is not available.
Notice that, in this across-regimes independence case, it makes sense to have $\pi_{ii}\neq 0$.
The $\alpha$ recursion is given by\footnote{The initialisation is given by $\alpha^{\sigma_1}_1=p(v_1|s_1)\tilde\pi_{s_1}p(\textrm{duration}\geq c_1)=p(v_1|s_1)\tilde\pi_{s_1}\prod_{i=1}^{c_1-1}\rho_i$. Notice that this way we are not imposing that the first regime starts at the first time-step $t=1$.}
\begin{align*}
\alpha^{\sigma_t}_t&=p(v_t|\sigma_t,\cancel{v_{1:t-k-1}},v_{t-k:t-1})p(\sigma_{t},v_{1:t-1})\\[3pt]
&=p(v_t|\sigma_t,v_{t-k:t-1})\sum_{\sigma_{t-1}}p(\sigma_t|\sigma_{t-1},\cancel{v_{1:t-1}})\alpha^{\sigma_{t-1}}_{t-1}\\[-3pt]
&=p(v_t|\sigma_t,v_{t-k:t-1})\Big\{\delta_{c_t>1}\lambda_{c_t-1}\alpha^{s_t,c_t-1}_{t-1}+\delta_{c_t=1}\hspace{-0.2cm}\sum_{c_{t-1}
=d_{\min}}^{d_{\max}}\hspace{-0.3cm}(1\!-\!\lambda_{c_{t-1}})\sum_{s_{t-1}}
\pi_{s_ts_{t-1}}\alpha^{\sigma_{t-1}}_{t-1}\Big\}\, ,
\end{align*}
where $p(v_t|\sigma_t,v_{t-k:t-1})=p(v_t|\sigma_t,v_{t-c_t+1:t-1})$ if $k\geq c_t$. The computational cost of this recursion is given by $O(TSd_{\max})$, namely the cost of the $\beta$ recursion in the previous model.

The $\beta$ recursion is obtained as follows\footnote{The initialisation is given by $\beta^{\sigma_T}_T=1$.}
\begin{align*}
\beta^{\sigma_t}_t&=\sum_{\sigma_{t+1}}p(v_{t+1:T}|\cancel{\sigma_t},\sigma_{t+1},v_{t-k+1:t})p(\sigma_{t+1}|\sigma_t,\cancel{v_{t-k+1:t}})\\
&=\sum_{\sigma_{t+1}}p(v_{t+2:T}|\sigma_{t+1},\cancel{v_{t-k+1}},v_{t-k+2:t+1})p(v_{t+1}|\sigma_{t+1},v_{t-k+1:t})p(\sigma_{t+1}|\sigma_t)\\[-7pt]
&=(1-\lambda_{c_t})\sum_{s_{t+1}}\!p(v_{t+1}|s_{t+1},c_{t+1}=1,v_{t-k+1:t})\pi_{s_{t+1}s_t}\beta^{s_{t+1},1}_{t+1}\\
&+\delta_{c_t<d_{\max}}\lambda_{c_t}p(v_{t+1}|s_t,c_{t+1}=c_t+1,v_{t-k+1:t})\beta^{s_t,c_t+1}_{t+1}\, .
\end{align*}
When pre-computing $\sum_{s_{t+1}}\!p(v_{t+1}|s_{t+1},c_{t+1}=1,v_{t-k+1:t})\pi_{s_{t+1}s_t}\beta^{s_{t+1},1}_{t+1}$, the computational cost of this recursion is given by $O(TS(S+d_{\max}))$, namely the cost of
the $\alpha$ recursion in the previous model.

\begin{figure}[t]
\renewcommand{\subfigcapskip}{0pt}
\begin{center}
\subfigure[]{\scalebox{0.7}{
\begin{tikzpicture}
\node[] at (1,1.25) {$\cdots$};
\node[disc] (sigmatm) at (2,1.25) {$c_{t-1}$};
\node[disc] (sigmat) at (4,1.25) {$c_t$};
\node[disc] (sigmatp) at (6,1.25) {$c_{t+1}$};
\node[disc] (dtm) at (2,0) {$d_{t-1}$};
\node[disc] (dt) at (4,0) {$d_t$};
\node[disc] (dtp) at (6,0) {$d_{t+1}$};
\node[disc] (stm) at (2,-1.25) {$s_{t-1}$};
\node[disc] (st) at (4,-1.25) {$s_t$};
\node[disc] (stp) at (6,-1.25) {$s_{t+1}$};
\node[] at (7,1.25) {$\cdots$};
\node[ocont,obs] (vtm) at (2,-2.5) {$v_{t-1}$};
\node[ocont,obs] (vt) at (4,-2.5) {$v_t$};
\node[ocont,obs] (vtp) at (6,-2.5) {$v_{t+1}$};
\node [red,above] at (sigmatm.north) {$c_{t-1}=2$};
\node [red,above] at (sigmat.north) {$c_t=1$};
\node [red,above] at (sigmatp.north) {$c_{t+1}>1$};
\draw[->, line width=1.15pt](stm)--(st);\draw[->, line width=1.15pt](st)--(stp);
\draw[->, line width=1.15pt](stm)--(vtm);\draw[->, line width=1.15pt](st)--(vt);\draw[->, line width=1.15pt](stp)--(vtp);
\draw[->, line width=1.15pt](dtm)--(dt);\draw[->, line width=1.15pt](dt)--(dtp);
\draw[->, line width=1.15pt](sigmatm)--(dt);\draw[->, line width=1.15pt](sigmat)--(dtp);
\draw[->, line width=1.15pt](sigmatm)--(st);\draw[->, line width=1.15pt](sigmat)--(stp);
\draw[->, line width=1.15pt](dtm)--(sigmatm);\draw[->, line width=1.15pt](dt)--(sigmat);\draw[->, line width=1.15pt](dtp)--(sigmatp);
\draw[->, line width=1.15pt](sigmatm)--(sigmat);\draw[->, line width=1.15pt](sigmat)--(sigmatp);
\draw[->, line width=1.15pt](sigmatm)to [bend left=35](vtm);\draw[->, line width=1.15pt](sigmat)to [bend left=35](vt);\draw[->, line width=1.15pt](sigmatp)to [bend left=35](vtp);
\draw[->, line width=1.15pt](dtm)to [bend right=45](vtm);\draw[->, line width=1.15pt](dt)to [bend right=45](vt);\draw[->, line width=1.15pt](dtp)to [bend right=45](vtp);
\draw[line width=1.15pt](vtm)--(vt);\draw[red,dashed,line width=1.15pt](vtm)to [bend left=35](vtp);\draw[red,dashed,line width=1.15pt](vt)--(vtp);
\end{tikzpicture}}}
\hspace{0.5cm}
\subfigure[]{\scalebox{0.7}{
\begin{tikzpicture}
\node[] at (1,1.25) {$\cdots$};
\node[disc] (sigmatm) at (2,1.25) {$c_{t-1}$};
\node[disc] (sigmat) at (4,1.25) {$c_t$};
\node[disc] (sigmatp) at (6,1.25) {$c_{t+1}$};
\node[disc] (dtm) at (2,0) {$d_{t-1}$};
\node[disc] (dt) at (4,0) {$d_t$};
\node[disc] (dtp) at (6,0) {$d_{t+1}$};
\node[disc] (stm) at (2,-1.25) {$s_{t-1}$};
\node[disc] (st) at (4,-1.25) {$s_t$};
\node[disc] (stp) at (6,-1.25) {$s_{t+1}$};
\node[] at (7,1.25) {$\cdots$};
\node[ocont,obs] (vtm) at (2,-2.5) {$v_{t-1}$};
\node[ocont,obs] (vt) at (4,-2.5) {$v_t$};
\node[ocont,obs] (vtp) at (6,-2.5) {$v_{t+1}$};
\node [red,above] at (sigmatm.north) {$c_{t-1}=2$};
\node [red,above] at (sigmat.north) {$c_t=1$};
\node [red,above] at (sigmatp.north) {$c_{t+1}>1$};
\draw[->, line width=1.15pt](stm)--(st);\draw[->, line width=1.15pt](st)--(stp);
\draw[->, line width=1.15pt](stm)--(vtm);\draw[->, line width=1.15pt](st)--(vt);\draw[->, line width=1.15pt](stp)--(vtp);
\draw[->, line width=1.15pt](dtm)--(dt);\draw[->, line width=1.15pt](dt)--(dtp);
\draw[->, line width=1.15pt](sigmatm)--(dt);\draw[->, line width=1.15pt](sigmat)--(dtp);
\draw[->, line width=1.15pt](sigmatm)--(st);\draw[->, line width=1.15pt](sigmat)--(stp);
\draw[->, line width=1.15pt](dtm)--(sigmatm);\draw[->, line width=1.15pt](dt)--(sigmat);\draw[->, line width=1.15pt](dtp)--(sigmatp);
\draw[->, line width=1.15pt](sigmatm)--(sigmat);\draw[->, line width=1.15pt](sigmat)--(sigmatp);
\draw[->, line width=1.15pt](sigmatm)to [bend left=35](vtm);\draw[->, line width=1.15pt](sigmat)to [bend left=35](vt);\draw[->, line width=1.15pt](sigmatp)to [bend left=35](vtp);
\draw[->, line width=1.15pt](dtm)to [bend right=45](vtm);\draw[->, line width=1.15pt](dt)to [bend right=45](vt);\draw[->, line width=1.15pt](dtp)to [bend right=45](vtp);
\draw[line width=1.15pt](vtm)--(vt);\draw[->, line width=1.15pt](vt)--(vtp);
\end{tikzpicture}}}
\end{center}
\caption{\HMSMs~in which the regime-duration distribution is explicitly modelled using two sets of discrete variables $c_{1:T},d_{1:T}$ as in \secref{sec:M2}.
In (a) the visible dependence from time $1,\ldots,t$ to time $t+1,\ldots,T$ is cut (red dashed lines), since $c_t=1$ indicates the end of the regime at time $t$. 
}
\label{fig:M2}
\end{figure}
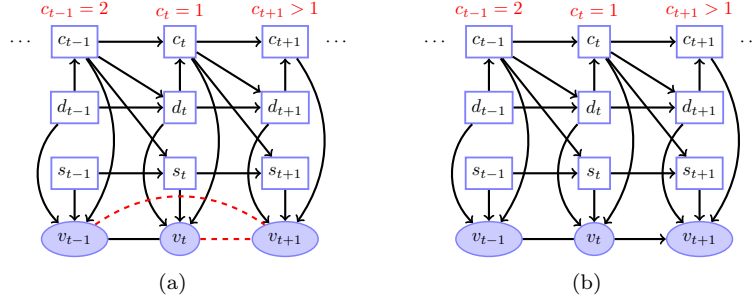

\subsection{Modelling the Regime-Duration Distribution with Two Sets of Duration-Count Variables \label{sec:M2}}
In this section, we describe how to model the regime-duration distribution using two separate sets of discrete variables $d_{1:T}$ and $c_{1:T}$. The duration variable $d_t$ specifies the number of time-steps spanned by the observations forming the current regime, and takes a value sampled from a regime-duration distribution. The count variable $c_t$ is modelled similarly to \secref{sec:M1st}, with the only difference that, at the beginning of a regime $t$, $c_t=d_t$ instead of being sampled from a regime-duration distribution. More specifically, by using the notation $\sigma_t=\{s_t,d_t,c_t\}$, we define $p(\sigma_t|\sigma_{t-1})=p(c_t|d_t,c_{t-1})p(d_t|d_{t-1},c_{t-1})p(s_t|s_{t-1},c_{t-1})$ with\footnote{For $t=1$, $p(s_1)=\tilde{\pi}_{s_1}, p(d_1)=\rho_{d_1}, p(c_1|d_1)\!=\!\delta_{c_1=d_1}$.}

\begin{minipage}{0.45\textwidth}
\begin{align*}
p(d_t|d_{t-1},c_{t-1})&=\begin{cases}
	\rho_{d_t} 			 \hspace{0.0cm}     &\textrm{if } c_{t-1}\!=\!1,\\
	\delta_{d_t=d_{t-1}} \hspace{0.00cm}	& \hspace{-0.0cm}\textrm{if } c_{t-1}\!>\!1,
\end{cases}
\end{align*}
\end{minipage}
\hspace{0.5cm}
\begin{minipage}{0.45\textwidth}
\begin{align*}
p(c_t|d_t,c_{t-1})&=\begin{cases}
	\delta_{c_t=d_t} 		& \textrm{if } c_{t-1}\!=\!1,\\
	\delta_{c_t=c_{t-1}-1} 	& \textrm{if } c_{t-1}\!>\!1,
\end{cases}
\end{align*}
\end{minipage}

and $p(s_t|s_{t-1},c_{t-1})$ as in \secref{sec:M1st}\footnote{In this model it makes sense to have $\pi_{ii}\neq 0$ even when a minimum regime duration is imposed.}.

The variables $d_t$ and $c_t$ provide information about the time-steps in the past and future at which the current
regime starts and ends. Therefore, unlike the previous models, this model can be used in the case of a
non-Markovian visible dependence within a regime. This is expressed by the undirected links among the observations
in the graphical model of \figref{fig:M2} (a). Notice that, both visible independence across regimes (\figref{fig:M2} (a))
or Markovian dependence across regimes (\figref{fig:M2} (b)) can be assumed. In the first case, these models are usually called segmental \HMMs.

In the sequel, we provide inference routines for the case in which independence is assumed.

\subsection*{Smoothing}
In order to perform smoothing, we first compute terms of the type $\gamma^{s_t,c_t,1}_t=p(s_t,d_t,c_t=1|v_{1:T})$ for which the count variable takes value 1.
If we define $\sigma^{1}_t\!=\!\{s_t,d_t,c_t\!=\!1\}$, we obtain\footnote{The normalisation term $p(v_{1:T})$ can be computed by summing \eqref{eq:gpost} over $s_t$ for a specific $t$,
or as $\sum_{s_T,d_T}\alpha^{s_T,d_T,1}_T$ when imposing the contraint $c_T=1$.}
\begin{align*}
\gamma^{s_t,d_t,1}_t\propto \underbrace{p(v_{t+1:T}|s_t,c_t\!=\!1,\cancel{d_t,v_{1:t}})}_{\beta^{s_t,1}_t}\underbrace{p(\sigma^{1}_t,v_{1:t})}_{\alpha^{s_t,d_t,1}_t}.
\end{align*}
The term $\alpha^{s_t,d_t,1}_t$ is estimated as follows\footnote{The initial and final recursions are as follows
\begin{align*}
&\textrm{for }t=1,\ldots,d_{\min}\hspace{0.2cm}\alpha^{s_t,d_t,1}_t=\left\{ \begin{array}{l}
p(v_{1:t}|\sigma^1_t)\tilde{\pi}_{s_t}\rho_{d_t}\textrm{ if } d_t=t\\
0\textrm{ if } d_t>t \textrm{ \& constraints } c_0=1, d_{\min}=1\\
p(v_{1:t}|\sigma^1_t)\tilde{\pi}_{s_t}\rho_{d_t}\textrm{ if } d_t>t \textrm{ \& no constraints } c_0=1, d_{\min}=1\\
\end{array}\right.\\
&\textrm{for }t=d_{\min}+1,\ldots,d_{\max}\hspace{0.2cm}
\alpha^{s_t,d_t,1}_t=
\left\{ \begin{array}{l}
p(v_{1:t}|\sigma^1_t)\tilde{\pi}_{s_t}\rho_{d_t}\textrm{ if } d_t=t\\
0\textrm{ if } d_t>t \textrm{ \& constraints } c_0=1, d_{\min}=1\\
p(v_{1:t}|\sigma^1_t)\tilde{\pi}_{s_t}\rho_{d_t}\textrm{ if } d_t>t \textrm{ \& no constraints } c_0=1, d_{\min}=1\\
p(v_{t-d_t+1:t}|\sigma^{1}_t)\rho_{d_t}\sum_{s_{t-d_t}}\pi_{s_ts_{t-d_t}}\sum_{d_{t-d_t}}\alpha^{s_{t-d_t},d_{t-d_t},1}_{t-d_t} \textrm{ if } d_t< t
\end{array}\right.\\[5pt]
&\textrm{for } t=d_{\max}+1,\ldots,T\hspace{0.2cm}\alpha^{s_t,d_t,1}_t=p(v_{t-d_t+1:t}|\sigma^{1}_t)\rho_{d_t}\sum_{s_{t-d_t}}\pi_{s_ts_{t-d_t}}\sum_{d_{t-d_t}}\alpha^{s_{t-d_t},d_{t-d_t},1}_{t-d_t}\\
&\textrm{for } t=T+1,\ldots,T+d_{\max}-1, \textrm{for } d_t=\max(d_{\min},t-T+1),\ldots,d_{\max}\hspace{0.2cm}\\
&\alpha^{s_t,d_t,1}_t=p(v_{t-d_t+1:T}|\sigma^{1}_t)\rho_{d_t}\sum_{s_{t-d_t}}\pi_{s_ts_{t-d_t}}\sum_{d_{t-d_t}}\alpha^{s_{t-d_t},d_{t-d_t},1}_{t-d_t}
\end{align*}
and $\beta^{j,1}_T=1$. Under the constraint $c_0=1,c_t=1$ with $d_{\min}>1$, extra care to ensure consistency at the beginning and end of the sequence is necessary. The case $c_0\neq 1,c_T\neq 1$ requires the definition of additional distribution for partial-regime observations.},

\hspace{0.1cm}
\vspace{-0.3cm}
\begin{align*}
\vspace{-3cm}\alpha^{s_t,d_t,1}_t&=p(v_{t-d_t+1:t}|\sigma^{1}_t,\cancel{v_{1:t-d_t}})p(\sigma^{1}_t,v_{1:t-d_t})\\[5pt]
&=p(v_{t-d_t+1:t}|\sigma^{1}_t)\hspace{-0.4cm}\sum_{s_{t-d_t},d_{t-d_t}}\hspace{-0.4cm}p(\sigma^{1}_t|\sigma^{1}_{t-d_t},\cancel{v_{1:t-d_t}})\alpha^{s_{t-d_t},d_{t-d_t},1}_{t-d_t}\\[-1pt]
&=p(v_{t-d_t+1:t}|\sigma^{1}_t)\rho_{d_t}\sum_{s_{t-d_t}}\pi_{s_ts_{t-d_t}}\underbrace{\sum_{d_{t-d_t}}\alpha^{s_{t-d_t},d_{t-d_t},1}_{t-d_t}}_{\hat\alpha^{s_{t-d_t},1}_{t-d_t}}
\end{align*}
Whilst the complexity would appear to be $O(TS^2(d_{\max}-d_{\min}+1)^2)$, by pre-computing $\hat\alpha^{s_{t-d_t},1}_{t-d_t}$ and $\tilde\alpha^{s_t,1}_{t-d_t}=\sum_{s_{t-d_t}}\pi_{s_ts_{t-d_t}}\hat\alpha^{s_{t-d_t},1}_{t-d_t}$, we obtain a complexity of
$O(TS(S+d_{\max}-d_{\min}+1))$ (without considering the complexity required to compute $q(v_{t-d_t+1:t}|\sigma^{1}_t)$).
Notice that this complexity is smaller than $O(TS^2d_{\max}^2)$ and $O(TS^2d_{\max})$ (case $d_{\min}=1$) stated in \cite{mitchell95complexity,murphy2002hsm,rabiner89tutorial,yu03efficient} for the same approach.

With the notation $\sigma^{k,1}_{t+k}=\{s_{t+k},d_{t+k}=k,c_{t+k}=1\}$, the term $\beta^{s_t,1}_t$ can be estimated with the following recursion
\begin{align*}
\beta^{s_t,1}_t&=\!\sum_{s_{t+k},k}\!p(v_{t+1:T}|\sigma^{k,1}_{t+k},s_t,c_t\!=\!1)p(\sigma^{k,1}_{t+k}|s_t,c_t\!=\!1)\\[-1pt]
&=\!\sum_{s_{t+k},k}\!p(v_{t+1:t+k}|\sigma^{k,1}_{t+k})\beta^{s_{t+k},1}_{t+k}\pi_{s_{t+k}s_t}\rho_k,
\end{align*}
which, by pre-computing  $p(v_{t+1:t+k}|\sigma^{1,k}_{t+k})\beta^{s_{t+k},1}_{t+k}\rho_k$, has complexity $O(TS(S+d_{\max}-d_{\min}+1))$. Therefore,
the $\alpha-\beta$ routines with separate duration-count variables
are computationally convenient over common duration-count variables
when $d_{\max}-d_{\min}+1\ll d_{\max}$.

Distributions of interest such as $p(s_t,c_t|v_{1:T})$ can be estimated as
\begin{align}
&p(s_t,c_t|v_{1:T})\propto\sum_{d=\max(d_{\min},c_t)}^{d_{\max}}\beta^{s_t,1}_{t+c_t-1}\alpha^{s_t,d,1}_{t+c_t-1}\nonumber\\
\label{eq:gpost}
&p(s_t|v_{1:T})\propto\sum_{\tau=t}^{t+d_{\max}-1}\sum_{d=\max(d_{\min},\tau-t+1)}^{d_{\max}}\beta^{s_t,1}_{\tau}\alpha^{s_t,d,1}_{\tau}\, .
\end{align}
The computational cost for $p(s_t,c_t|v_{1:T})$ is given by $O(TSd_{\min}(d_{\max}-d_{\min}+1)+(d_{\max}-d_{\min})(d_{\max}-d_{\min}+1)/2)$.
Indeed for $c_t=1,\ldots,d_{\min}$ we have $d_{\max}-d_{\min}+1$ summations to perform,
whilst for $c_t=d_{\min}+1,\ldots,d_{\max}$ we have $d_{\max}-c_t+1$ summations to perform,
which gives rise to $1+2+\ldots+d_{\max}-d_{\min}=(d_{\max}-d_{\min})(d_{\max}-d_{\min}+1)/2$.

\paragraph{Case $\pi_{ii}=0$.} 
In this case, it is more convenient to compute $p(s_t|v_{1:T})$, using the recursion for
$\alpha^{s_t,:,1}_t=\sum_{d_t}p(v_{t-d_t+1:t}|\sigma^{1}_t)\rho_{d_t}\tilde\alpha^{s_t,:,1}_{t-d_t}$
which has complexity $O(TS(S+d_{\max}-d_{\min}+1))$, and then calculate
\begin{align*}
&p(s_t|v_{1:T})=\sum_{\tau<t}^{}\beta^{*,s_t,1}_{\tau}\alpha^{*,s_t,:,1}_{\tau}-\beta^{s_t,1}_{\tau}\alpha^{s_t,:,1}_{\tau}\, .
\end{align*}


\subsection*{Most Likely Sequence of Hidden Variables}
If we define $\delta^{s_t,d_t,1}_t=\max_{s_{1:t-1},d_{1:t-1}}p(s_{1:t-1},d_{1:t-1},\sigma^{1}_t,v_{1:t})$,
then the most likely sequence of states, duration and count variables $s^*_{1:T},d^*_{1:T},c^*_{1:T}$ (under the assumptions $c_T=1$) can be obtained as
\begin{align*}
&\textrm{for } t=1,\ldots,d_{\min}\hspace{0.2cm}\delta^{s_t,d_t,1}_t=\alpha^{s_t,d_t,1}_t\\[5pt]
&\hspace{-0.1cm}\left.
\textrm{for } t=d_{\min}+1,\ldots,d_{\max}
\right.\hspace{0.2cm}
\delta^{s_t,d_t,1}_t=
\left\{ \begin{array}{l}
\alpha^{s_t,d_t,1}_t\hspace{0.0cm} \textrm{ if } d_t\geq t\\ 
p(v_{t-d_t+1:t}|\sigma^{1}_t)\rho_{d_t}\max_{s_{t-d_t}}\pi_{s_ts_{t-d_t}}\max_{d_{t-d_t}}\delta^{s_{t-d_t},d_{t-d_t},1}_{t-d_t}\textrm{ if } d_t< t\\
\end{array}\right.\\[5pt]
&\textrm{for } t=d_{\max}+1,\ldots,T\hspace{0.2cm}
\left.\begin{array}{l}
\delta^{s_t,d_t,1}_t=p(v_{t-d_t+1:t}|\sigma^{1}_t)\rho_{d_t}\max_{s_{t-d_t}}\pi_{s_ts_{t-d_t}}\max_{d_{t-d_t}}\delta^{s_{t-d_t},d_{t-d_t},1}_{t-d_t}\\
\psi^{s_t,d_t}_t=\argmax_{s_{t-d_t},d_{t-d_t}}\pi_{s_ts_{t-d_t}}\delta^{s_{t-d_t},d_{t-d_t},1}_{t-d_t}
\end{array}\right.\\[5pt]
&[s^*_T,d^*_T,c^*_T]=\left\{ \begin{array}{l}
\{\argmax_{s_T,d_T} \delta^{s_T,d_T,1}_T, 1\} \textrm{ if contraint } c_T=1\\
\argmax_{s_T,d_T,c_T} \alpha^{s_T,d_T,c_T}_T \textrm{ if no contraint } c_T=1\\
\end{array}\right.\\[5pt]
&s^*_{T-d^*_T+c^*_T:T-1}=s^*_T, d^*_{T-d^*_T+c^*_T:T-1}=d^*_T, c^*_{T-d^*_T+c^*_T:T-1}=d^*_T,\ldots,c^*_T+1\\
&t=T-d^*_T+c^*_T-1,\hspace{0.1cm} \textrm{while } t>1\hspace{0.2cm}
\left.\begin{array}{l}
\{s^*_t,d^*_t,c^*_t\}=\{\psi^{s^*_{t+1},d^*_{t+1}}_{t+1},1\}\\
s^*_{t-d^*_t+1:t-1}=s^*_t,\hspace{0.2cm} d^*_{t-d^*_t+1:t-1}=d^*_t,\hspace{0.2cm} c^*_{t-d^*_t+1:t-1}=d^*_t,\ldots,2\\
t=t-d^*_t
\end{array}\right.
\end{align*}

\subsection*{Sampling a Sequence of Hidden Variables from $p(\sigma_{1:T}|v_{1:T})$}
Such samples can be obtained recursively by considering the factorisation $p(\sigma_{1:T}|v_{1:T})=p(\sigma_T|v_{1:T})\prod_{t=1}^{T-1}p(\sigma_t|\sigma_{t+1},v_{1:T})$.
Suppose that, at time $t$, $c_t=1$ and we have sampled regime type $s_t$ and a duration $d_t$.
Then, $s_{t-d_t+1:t-1}=s_t$, $d_{t-d_t+1:t-1}=d_t$ and $c_{t-d_t+1:t-1}=d_t,\ldots,2, c_{t-k}=1$, so that effectively we
need to sample $s_{t-d_t},d_{t-d_t}$ from the distribution $p(\sigma^1_{t-d_t}|\sigma_{t-d_t+1},v_{1:t},\cancel{v_{t+1:T}})$, which is given by
\begin{align*}
p(\sigma^1_{t-d_t}|\sigma_{t-d_t+1},v_{1:t+1})&=\frac{p(\sigma^1_{t-d_t},\sigma_{t-d_t+1},v_{1:t})}{p(\sigma_{t-d_t+1},v_{1:t})}\\
&=\frac{\pi_{s_ts_{t-d_t}}\alpha^{s_{t-d_t},d_{t-k},1}_{t-d_t}}{\sum_i\pi_{s_ti}\sum_l\alpha^{i,l,1}_{t-d_t}}.
\end{align*}

\paragraph{Note on Complexity.} It is clear from the complexity analysis described above that whether to use an approach with joint or separate duration-count variables depends on the specific goals and assumptions. If we are interested in estimating the most likely sequence of hidden variables or sampling a path only, then if $d_{\max}-d_{\min}+1\ll d_{\max}$, using separate variables is more convenient.
However, when interested in distributions such as, then one has to take into account the extra cost incurred in computing these distributions.

\section{Switching Linear Gaussian State-Space Models}
In Sections \ref{sec:HMM_DO} and \ref{sec:M1M2}, we have analysed how to improve the basic \HMM~model by
enabling dependence among observations and by explicit modeling of the regime-duration distribution.
In this section, we introduce another extension of the \HMM~called the Switching Linear Gaussian State-Space Model (\SLGSSM),
which enables a more accurate modeling of noisy
and continuous observations and the reconstruction of
the hidden dynamics underlying a sequence of noisy observations
\cite{mesot-barber-07a,quinn08factorial,zoeter05ep}.

The \SLGSSM~is defined by the following linear equations
\begin{align*}
&h_t=A^{s_t}h_{t-1}+\eta^h_t,\hspace{0.2cm}\eta^h_t\sim{\cal N}(\eta^h_t;0,\Sigma^{s_t}_H),\hspace{0.05cm}h_1\sim{\cal N}(h_1;\mu^{s_t},\Sigma^{s_t}),\\[5pt]
&v_t=B^{s_t}h_t+\eta^v_t,\hspace{0.2cm}\eta^v_t\sim{\cal N}(\eta^v_t;0,\Sigma^{s_t}_V),
\end{align*}
where the continuous hidden variables $h_{1:T}$ represents a hidden dynamics. The joint distribution of all variables factorizes as follows
\begin{align*}
p(v_{1:T},h_{1:T},s_{1:T})=p(v_1|h_1,s_1)p(h_1|s_1)p(s_1)\prod_{t=2}^Tp(v_t|h_t,s_t)p(h_t|h_{t-1},s_t)p(s_t|s_{t-1}).
\end{align*}

Performing inference in the \SLGSSM~is intractable since, e.g., $p(h_t|s_t,v_{1:t})$ is a Gaussian mixture with $S^{t-1}$ components.
Several approximation schemes have been introduced over the years in order to deal with this intractability issue. For filtering,
a successful method consists of employing a Gaussian collapsing procedure in a filtering routine
on the line of the standard \LGSSM~predictor-corrector filtering routine. For smoothing,
\cite{barber06ec} showed how, by introducing some approximations, it is possible to
use Gaussian collapsing in a Rauch-Tung-Striebel style smoothing routine, obtaining improved accuracy over
other approximation schemes.

\section{Extension of \SLGSSM~with Explicit Regime-Duration Distribution \label{sec:SLGSSMM1M2}}
As in the classical \HMM, the regime-duration
distribution of the \SLGSSM~is implicitly geometric.
Therefore the \SLGSSM~shares the same limitations of the \HMM~in this respect.
In this section, we show how apply methods analogous to the ones
described in \secref{sec:M1M2} for explicitly modeling the regime-duration distribution of the \SLGSSM.

We will use a Gaussian collapsing approximation method on the line of \cite{barber06ec}.
Extensions of the \SLGSSM~for defining an explicit duration distribution have also been studied in \cite{oh08learning}.
The authors introduce a set of duration-count variables $c_{1:T}$ as in \secref{sec:M1st}, and use
standard approximate inference methods for the \SLGSSM~in which the discrete hidden variables
is a merged variable $\{c_t,s_t\}$. Sparsity in the transition distribution of the merged
variables is then used to reduce computational cost. The advantage of maintaining separate
variables is to get insight into the problem and properties of the approach.
As we will see, this will enables us to derive exact inference for the special case in which independence
among observations is imposed.

\subsection{Modeling State-Duration Distribution with One Set of Count-Duration Variables}
In this section, we describe how define an explicit state-duration distribution for the
\SLGSSM~by using one set of count-duration variables $c_{1:T}$ as in \secref{sec:M1}.

\subsubsection{Decreasing Duration-Count Variables \label{sec:SLGSSMM1st}}
\begin{figure}
\subfigure[]{\scalebox{0.7}{
\begin{tikzpicture}[dgraph]
\node[] at (1,0.2) {$\cdots$};
\node[disc] (sigmatm) at (2,0.2) {$c_{t-1}$};
\node[disc] (sigmat) at (4,0.2) {$c_t$};
\node[disc] (sigmatp) at (6,0.2) {$c_{t+1}$};
\node[] at (7,0.2) {$\cdots$};
\node[disc] (stm) at (2,-1.05) {$s_{t-1}$};
\node[disc] (st) at (4,-1.05) {$s_t$};
\node[disc] (stp) at (6,-1.05) {$s_{t+1}$};
\node[ocont] (htm) at (2,-2.25) {$h_{t-1}$};
\node[ocont] (ht) at (4,-2.25) {$h_t$};
\node[ocont] (htp) at (6,-2.25) {$h_{t+1}$};
\node[ocont,obs] (vtm) at (2,-3.5) {$v_{t-1}$};
\node[ocont,obs] (vt) at (4,-3.5) {$v_t$};
\node[ocont,obs] (vtp) at (6,-3.5) {$v_{t+1}$};
\draw[line width=1.15pt](sigmatm)--(sigmat);\draw[line width=1.15pt](sigmat)--(sigmatp);
\draw[line width=1.15pt](stm)--(st);\draw[line width=1.15pt](st)--(stp);
\draw[line width=1.15pt](stm)--(htm);\draw[line width=1.15pt](st)--(ht);\draw[line width=1.15pt](stp)--(htp);
\draw[line width=1.15pt](sigmatm)--(st);\draw[line width=1.15pt](sigmat)--(stp);
\draw[line width=1.15pt](htm)--(vtm);\draw[line width=1.15pt](ht)--(vt);\draw[line width=1.15pt](htp)--(vtp);
\draw[line width=1.15pt](htm)--(ht);\draw[line width=1.15pt](ht)--(htp);
\draw[line width=1.15pt](stm)to [bend right=55](vtm);\draw[line width=1.15pt](st)to [bend right=55](vt);\draw[line width=1.15pt](stp)to [bend right=55](vtp);
\end{tikzpicture}}}
\hspace{-0.4cm}
\subfigure[]{\scalebox{0.7}{
\begin{tikzpicture}[dgraph]
\node[] at (1,0.2) {$\cdots$};
\node[disc] (sigmatm) at (2,0.2) {$c_{t-1}$};
\node[disc] (sigmat) at (4,0.2) {$c_t$};
\node[disc] (sigmatp) at (6,0.2) {$c_{t+1}$};
\node[] at (7,0.2) {$\cdots$};
\node[disc] (stm) at (2,-1.05) {$s_{t-1}$};
\node[disc] (st) at (4,-1.05) {$s_t$};
\node[disc] (stp) at (6,-1.05) {$s_{t+1}$};
\node[ocont] (htm) at (2,-2.25) {$h_{t-1}$};
\node[ocont] (ht) at (4,-2.25) {$h_t$};
\node[ocont] (htp) at (6,-2.25) {$h_{t+1}$};
\node[ocont,obs] (vtm) at (2,-3.5) {$v_{t-1}$};
\node[ocont,obs] (vt) at (4,-3.5) {$v_t$};
\node[ocont,obs] (vtp) at (6,-3.5) {$v_{t+1}$};
\node [red,above] at (sigmatm.north) {$c_{t-1}=1$};
\draw[line width=1.15pt](sigmatm)--(ht);\draw[line width=1.15pt](sigmat)--(htp);
\draw[line width=1.15pt](sigmatm)--(sigmat);\draw[line width=1.15pt](sigmat)--(sigmatp);
\draw[line width=1.15pt](stm)--(st);\draw[line width=1.15pt](st)--(stp);
\draw[line width=1.15pt](stm)--(htm);\draw[line width=1.15pt](st)--(ht);\draw[line width=1.15pt](stp)--(htp);
\draw[line width=1.15pt](sigmatm)--(st);\draw[line width=1.15pt](sigmat)--(stp);
\draw[line width=1.15pt](htm)--(vtm);\draw[line width=1.15pt](ht)--(vt);\draw[line width=1.15pt](htp)--(vtp);
\draw[red,dashed,line width=1.15pt](htm)--(ht);\draw[line width=1.15pt](ht)--(htp);
\draw[line width=1.15pt](stm)to [bend right=55](vtm);\draw[line width=1.15pt](st)to [bend right=55](vt);\draw[line width=1.15pt](stp)to [bend right=55](vtp);
\end{tikzpicture}}}
\hspace{-0.4cm}
\subfigure[]{\scalebox{0.7}{
\begin{tikzpicture}[dgraph]
\node[] at (1,0.2) {$\cdots$};
\node[disc] (sigmatm) at (2,0.2) {$c_{t-1}$};
\node[disc] (sigmat) at (4,0.2) {$c_t$};
\node[disc] (sigmatp) at (6,0.2) {$c_{t+1}$};
\node[] at (7,0.2) {$\cdots$};
\node[disc] (stm) at (2,-1.05) {$s_{t-1}$};
\node[disc] (st) at (4,-1.05) {$s_t$};
\node[disc] (stp) at (6,-1.05) {$s_{t+1}$};
\node[ocont] (htm) at (2,-2.25) {$h_{t-1}$};
\node[ocont] (ht) at (4,-2.25) {$h_t$};
\node[ocont] (htp) at (6,-2.25) {$h_{t+1}$};
\node[ocont,obs] (vtm) at (2,-3.5) {$v_{t-1}$};
\node[ocont,obs] (vt) at (4,-3.5) {$v_t$};
\node[ocont,obs] (vtp) at (6,-3.5) {$v_{t+1}$};
\node [red,above] at (sigmat.north) {$c_{t}=1$};
\draw[line width=1.15pt](sigmatm)to [bend right=55](htm);\draw[line width=1.15pt](sigmat)to [bend right=55](ht);\draw[line width=1.15pt](sigmatp)to [bend right=55](htp);
\draw[line width=1.15pt](sigmatm)--(sigmat);\draw[line width=1.15pt](sigmat)--(sigmatp);
\draw[line width=1.15pt](stm)--(st);\draw[line width=1.15pt](st)--(stp);
\draw[line width=1.15pt](stm)--(htm);\draw[line width=1.15pt](st)--(ht);\draw[line width=1.15pt](stp)--(htp);
\draw[line width=1.15pt](sigmatm)--(stm);\draw[line width=1.15pt](sigmat)--(st);\draw[line width=1.15pt](sigmat)--(st);
\draw[line width=1.15pt](htm)--(vtm);\draw[line width=1.15pt](ht)--(vt);\draw[line width=1.15pt](htp)--(vtp);
\draw[red,dashed,line width=1.15pt](htm)--(ht);\draw[line width=1.15pt](ht)--(htp);
\draw[line width=1.15pt](stm)to [bend right=55](vtm);\draw[line width=1.15pt](st)to [bend right=55](vt);\draw[line width=1.15pt](stp)to [bend right=55](vtp);
\end{tikzpicture}}}
\caption{Extension of a sLGDS with duration-count variables.}
\label{fig:SLGSS}
\vspace{-0cm}
\end{figure}
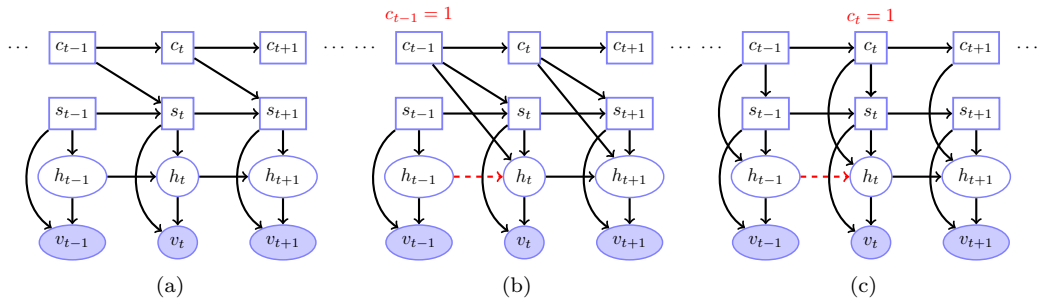

The first approach is by modeling the duration-count variables as in \secref{sec:M1st}, obtaining the belief network representation in \figref{fig:SLGSS} (a). 
In the sequel, we will describe how to perform smoothing in this model. Unlike the models explained in the previous sections,
this will be achieved with a sequential approach that first computes the filtered distributions $p(h_t,s_t|v_{1:t})$ and
then uses this estimate to compute $p(h_t,s_t|v_{1:T})$. This approach is preferable, since working in a log scale is not possible.
\subsection*{Filtering}
An approach to filtering is to form the following recursion for $p(h_t|\sigma_t,v_{1:t})$
\begin{align}
p(h_t|\sigma_t,v_{1:t})&=\sum_{\sigma_{t-1}}p(h_t|\sigma_{t-1},s_t,\cancel{c_t},v_{1:t})p(\sigma_{t-1}|\sigma_t,v_{1:t})\nonumber\\
&=\delta_{c_t<d_{\max}}p(h_t|s_{t-1}=s_t,c_{t-1}=c_t+1,s_t,v_{1:t})p(s_{t-1}=s_t,c_{t-1}=c_t+1|\sigma_t,v_{1:t})\nonumber\\
&+\sum_{s_{t-1}}p(h_t|s_{t-1},c_{t-1}=1,s_t,v_{1:t})p(s_{t-1},c_{t-1}=1|\sigma_t,v_{1:t})\, .
\label{eq:filt}
\end{align}
In order to understand the complexity of this recursion, suppose for simplicity that $d_{\min}=1$.
Then $p(h_1|\sigma_1,v_1)$ is a Gaussian and, from the recursion\footnote{We use the fact if $p(h_{t-1}|\sigma_{t-1},v_{1:t-1})$ is a mixture of Gaussians, then $p(h_t|\sigma_{t-1},s_t,v_{1:t})$ is also a mixture of Gaussians with the same number of components
(see below).}, we see that $p(h_2|\sigma_2,v_{1:2})$ is a mixture of Gaussian with $S$ components. Therefore, in general, $p(h_t|\sigma_t,v_{1:t})$ is a mixture of Gaussians with $S^{t-1}$ components as in the standard \SLGSSM. In order to overcome, this exponential
explosion of component with time, after estimating the recursion, we collapse the obtained mixture to a single Gaussian.

Below we describe how to compute each required term in the recursion.
\paragraph*{Computing $p(h_t|\sigma_{t-1},s_t,v_{1:t})$.}
We have
\begin{align*}
p(h_t|\sigma_{t-1},s_t,v_{1:t})
&=\frac{p(v_t,h_t|\sigma_{t-1},s_t,v_{1:t-1})}{p(v_t|\sigma_{t-1},s_t,v_{1:t-1})}\\
&=\frac{p(v_t|\cancel{\sigma_{t-1}},s_t,h_t,\cancel{v_{1:t-1}})p(h_t|\sigma_{t-1},s_t,v_{1:t-1})}{p(v_t|\sigma_{t-1},s_t,v_{1:t-1})}\\
&=\frac{p(v_t|s_t,h_t)\int_{h_{t-1}}p(h_{t-1:t}|\sigma_{t-1},s_t,v_{1:t-1})}{p(v_t|\sigma_{t-1},s_t,v_{1:t-1})}\\
&=\frac{p(v_t|s_t,h_t)\int_{h_{t-1}}p(h_t|h_{t-1},\cancel{\sigma_{t-1}},s_t,\cancel{v_{1:t-1}})p(h_{t-1}|\sigma_{t-1},\cancel{s_t},v_{1:t-1})}{p(v_t|\sigma_{t-1},s_t,v_{1:t-1})}.
\end{align*}
If we assume that $p(h_{t-1}|\sigma_{t-1},v_{1:t-1})$ is a Gaussian distribution with mean $\hat h_{t-1}^{t-1,\sigma_{t-1}}$ and covariance $P_{t-1}^{t-1,\sigma_{t-1}}$, by using the linear system equations defining the evolution of the continuous hidden variables we find that $p(h_t|\sigma_{t-1},s_t,v_{1:t-1})$ is Gaussian with mean and covariance given by
\begin{align*}
\hat h_t^{t-1,\sigma_{t-1},s_t}&=\av{h_t}_{p(h_t|\sigma_{t-1},s_t,v_{1:t-1})}=A^{s_t}\hat h_{t-1}^{t-1,\sigma_{t-1}}\\[7pt]
P_t^{t-1,\sigma_{t-1},s_t}&=\av{(h_t-h_t^{t-1,\sigma_{t-1},s_t})(h_t-h_t^{t-1,\sigma_{t-1},s_t})\trans}_{p(h_t|\sigma_{t-1},s_t,v_{1:t-1})}\\[3pt]
&=A^{s_t}P_{t-1}^{t-1,\sigma_{t-1}}(A^{s_t})\trans+\Sigma^{s_t}_H.
\end{align*}

By using the linear system equations defining the observation-generation process we find
\begin{align*}
&\av{v_t}_{p(v_t|\sigma_{t-1},s_t,v_{1:t-1})}=B^{s_t}\hat h_t^{t-1,\sigma_{t-1},s_t}\\[5pt]
&\av{(v_t-\av{v_t})(v_t-\av{v_t})\trans}_{p(v_t|\sigma_{t-1},s_t,v_{1:t-1})}=B^{s_t}P_t^{t-1,\sigma_{t-1},s_t}(B^{s_t})\trans+\Sigma^{s_t}_V\\[5pt]
&\av{(v_t-\av{v_t})(h_t-\hat h_t^{t-1,\sigma_{t-1},s_t})\trans}_{p(v_t,h_t|\sigma_{t-1},s_t,v_{1:t-1})}=B^{s_t}P_{t}^{t-1,\sigma_{t-1},s_t}.
\end{align*}
%

By using the formula of Gaussian conditioning, we find that $p(h_t|\sigma_{t-1},s_t,v_{1:t})$ is Gaussian with mean and covariance given by
\begin{align*}
&\hat h_t^{t,\sigma_{t-1},s_t}=\hat h_t^{t-1,\sigma_{t-1},s_t}+K(v_t-B^{s_t}\hat h_t^{t-1,\sigma_{t-1},s_t})\\[5pt]
&P_t^{t,\sigma_{t-1},s_t}=P_t^{t-1,\sigma_{t-1},s_t}-KB^{s_t}P_t^{t-1,s_{t-1},s_t,c_{t-1}}\, ,
\end{align*}
where $K=P_t^{t-1,\sigma_{t-1},s_t}(B^{s_t})\trans(B^{s_t}P_t^{t-1,\sigma_{t-1},s_t}(B^{s_t})\trans+\Sigma^{s_t}_V)^{-1}$.
\paragraph*{Computing $p(\sigma_{t-1}|\sigma_t,v_{1:t})$.} We have
\begin{align*}
p(\sigma_t|v_{1:t})&=\frac{p(v_t,\sigma_t|v_{1:t-1})}{p(v_t|v_{1:t-1})}\\
&\propto\sum_{\sigma_{t-1}}p(v_t,\sigma_{t-1:t}|v_{1:t-1})\\
&=\sum_{\sigma_{t-1}} p(v_t|\sigma_{t-1},s_t,v_{1:t-1})p(c_t|c_{t-1})p(s_t|\sigma_{t-1})p(\sigma_{t-1}|v_{1:t-1})\\
&=\delta_{c_t<d_{\max}}p(v_t|s_{t-1}=s_t,c_{t-1}=c_t+1,s_t,v_{1:t-1})p(s_{t-1}=s_t,c_{t-1}=c_t+1|v_{1:t-1})\\
&+\sum_{s_{t-1}}p(v_t|s_{t-1},c_{t-1}=1,s_t,v_{1:t-1})\rho_{c_t}\pi_{s_ts_{t-1}}p(s_{t-1},c_{t-1}=1|v_{1:t-1}),
\end{align*}
where $p(v_t|\sigma_{t-1},s_t,v_{1:t-1})=\frac{1}{\sqrt{\det P_{t}^{t-1,\sigma_{t-1},s_t}}}e^{-\frac{1}{2}(v_t-\hat h_t^{t-1,\sigma_{t-1},s_t})(P_{t}^{t-1,\sigma_{t-1},s_t})^{-1}(v_t-\hat h_t^{t-1,\sigma_{t-1},s_t})}$. Therefore
\begin{align}
p(\sigma_{t-1}|\sigma_t,v_{1:t})&=\frac{p(\sigma_{t-1:t}|v_{1:t})}{p(\sigma_t|v_{1:t})}\nonumber\\
&=\frac{p(v_t,\sigma_{t-1:t}|v_{1:t-1})}{p(v_t|v_{1:t-1})p(\sigma_{t}|v_{1:t})}\nonumber\\
&=\frac{p(v_t|\sigma_{t-1},s_t,v_{1:t-1})p(c_t|c_{t-1})p(s_t|\sigma_{t-1})
p(\sigma_{t-1}|v_{1:t-1})}{p(v_t|v_{1:t-1})p(\sigma_{t}|v_{1:t})}\, .
\label{eq:cond}
\end{align}

\paragraph*{Gaussian Collapsing.}
We can collapse the mixture of $S$ Gaussians $p(h_t|\sigma_t,v_{1:t})$ to a Gaussian by moment matching
\begin{align*}
\hat h_t^{t,\sigma_t}&=\delta_{c_t<d_{\max}}p(s_{t-1}=s_t,c_{t-1}=c_t+1|\sigma_t,v_{1:t})\hat h_t^{t,s_t,c_t+1,s_t}\\
&+\sum_{s_{t-1}}p(s_{t-1},c_{t-1}=1|\sigma_t,v_{1:t})\hat h_t^{t,s_{t-1},1,s_t}\\
P_t^{t,\sigma_t}&=p(s_{t-1}=s_t,c_{t-1}=c_t+1|\sigma_t,v_{1:t})(P_t^{t,s_t,c_t+1,s_t}+\hat h_t^{t,s_t,c_t+1,s_t}(\hat h_t^{t,s_t,c_t+1,s_t})\trans)\\
&+\sum_{s_{t-1}}p(s_{t-1},c_{t-1}=1|\sigma_t,v_{1:t})(P_t^{t,s_{t-1},1,s_t}+\hat h_t^{t,s_{t-1},1,s_t}(\hat h_t^{t,s_{t-1},1,s_t})\trans)-\hat h_t^{t,\sigma_t}(\hat h_t^{t,\sigma_t})\trans.
\end{align*}

The computational cost of this filtering recursion is $O(TS^2(d_{\max}-d_{\min}+1))$. However, notice that if
we plug \eqref{eq:cond} into \eqref{eq:filt} we obtain
\begin{align*}
&p(h_t|\sigma_t,v_{1:t})\\
&=\delta_{c_t<d_{\max}}p(h_t|s_{t-1}=s_t,c_{t-1}=c_t+1,s_t,v_{1:t})p(s_{t-1}=s_t,c_{t-1}=c_t+1|\sigma_t,v_{1:t})\nonumber\\
&+\frac{\rho_{c_t}}{p(v_t|v_{1:t-1})p(\sigma_{t}|v_{1:t})}
\sum_{s_{t-1}}p(h_t|s_{t-1},c_{t-1}=1,s_t,v_{1:t})p(v_t|s_{t-1},c_{t-1}=1,s_t,v_{1:t-1})\pi_{s_ts_{t-1}}p(\sigma_{t-1}|v_{1:t-1})
\end{align*}
Therefore one sum only over $s_{t-1}$ for all $c_t$ is required and the cost reduces to $O(T(S(d_{\max}-d_{\min}+1)+S))$.

\paragraph{Cut of dependence when changing regime.}
Notice that in this model, we can introduce cut of dependence when changing regime by adding a link from $c_{t-1}$ to $h_t$ as in \figref{fig:SLGSS} (b), and therefore in principle without the need to use a modeling for $c_{1:T}$ as in \secref{sec:M1alt}.
In this case
\begin{align*}
p(h_t|h_{t-1},s_t,c_{t-1})&=\begin{cases}
    {\cal N}(h_t;\mu^{s_t},\Sigma^{s_t}) \hspace{0.57cm}	& \textrm{if } c_{t-1}=1\\
	{\cal N}(h_t;A^{s_t}h_{t-1},\Sigma^{s_t}_H) 	& \textrm{if } c_{t-1}>1\\
\end{cases}
\end{align*}
Therefore, the filtering recursion becomes
\begin{align*}
p(h_t|\sigma_t,v_{1:t})&=\sum_{\sigma_{t-1}}p(h_t|\sigma_{t-1},s_t,\cancel{c_t},v_{1:t})p(\sigma_{t-1}|\sigma_t,v_{1:t})\\
&=\delta_{c_t<d_{\max}}p(h_t|s_{t-1}=s_t,c_{t-1}=c_t+1,s_t,v_{1:t})p(s_{t-1}=s_t,c_{t-1}=c_t+1|\sigma_t,v_{1:t})\\
&+\sum_{s_{t-1}}p(h_t|\cancel{s_{t-1}},c_{t-1}=1,s_t,\cancel{v_{1:t-1}},v_t)p(s_{t-1},c_{t-1}=1|\sigma_t,v_{1:t})\\
&=\delta_{c_t<d_{\max}}p(h_t|s_{t-1}=s_t,c_{t-1}=c_t+1,s_t,v_{1:t})p(s_{t-1}=s_t,c_{t-1}=c_t+1|\sigma_t,v_{1:t})\\
&+\frac{p(v_t|h_t,s_t){\cal N}(h_t;\mu^{s_t},\Sigma^{s_t})}{p(v_t|s_t,c_{t-1}=1)} \sum_{s_{t-1}}p(s_{t-1},c_{t-1}=1|\sigma_t,v_{1:t})\\
&=\delta_{c_t<d_{\max}}\frac{p(v_t|s_t,h_t)\int_{h_{t-1}}{\cal N}(h_t;A^{s_t}h_{t-1},\Sigma^{s_t}_H)
p(h_{t-1}|\sigma_{t-1},\cancel{s_t},v_{1:t-1})}{p(v_t|\sigma_{t-1},s_t,v_{1:t-1})}\\
&p(s_{t-1}=s_t,c_{t-1}=c_t+1|\sigma_t,v_{1:t})\\
&+\frac{p(v_t|h_t,s_t){\cal N}(h_t;\mu^{s_t},\Sigma^{s_t})}{p(v_t|s_t,c_{t-1}=1)}p(c_{t-1}\!=\!1|\sigma_t,v_{1:t}).
\end{align*}
From this recursion we see that $p(h_t|\sigma_t,v_{1:t})$ is a mixture of Gaussian with $t$ components (and $p(h_t|v_{1:t})$ is a mixture of Gaussian with $Sd_{\max}t$ components). It is therefore feasible to perform inference without the collapsing approximation. However, as we will see below, it is computationally more convenient to use a modeling for $c_{1:T}$ as in \secref{sec:M1alt}, for which $p(h_t|v_{1:t})$ is a mixture of Gaussian with $Sd_{\max}$ components.

\subsection*{Smoothing}
For smoothing we use the following recursion
\begin{align*}
p(h_t|\sigma_t,v_{1:T})&=\sum_{\sigma_{t+1}}p(h_t|\sigma_{t:t+1},v_{1:T})p(\sigma_{t+1}|\sigma_t,v_{1:T})\\
&=\delta_{c_t=1}\sum_{\sigma_{t+1}}p(h_t|\sigma_{t:t+1},v_{1:T})p(\sigma_{t+1}|\sigma_t,v_{1:T})\\
&+\delta_{c_t>1}p(h_t|\sigma_t,s_{t+1}=s_t,c_{t+1}=c_t-1,v_{1:T})\overbrace{p(s_{t+1}=s_t,c_{t+1}=c_t-1|\sigma_t,v_{1:T})}^{1}\,.
\end{align*}
Notice that in general $h_t \cancel{\ci} c_{t+1}\,|\, \sigma_t,s_{t+1},v_{1:T}$ since, for example, the path $c_{t+1},s_{t+2},v_{t+2},h_{t+2},h_{t+1},h_t$ is not blocked. However for $c_t>1$, $p(h_t|\sigma_t,s_{t+1}=s_t,c_{t+1}=c_t-1,v_{1:T})=
p(h_t|\sigma_t,v_{1:T})$.

\paragraph{Computing $p(h_t|\sigma_{t:t+1},v_{1:T})$.}
\begin{align}
p(h_t|\sigma_{t:t+1},v_{1:T})&=\int_{h_{t+1}}p(h_t|h_{t+1},\sigma_{t},s_{t+1},\cancel{c_{t+1}},v_{1:t},\cancel{v_{t+1:T}})
p(h_{t+1}|\sigma_{t:t+1}v_{1:T})\nonumber\\
&\approx \int_{h_{t+1}}p(h_t|h_{t+1},\sigma_{t},s_{t+1},v_{1:t})p(h_{t+1}|\sigma_{t+1},v_{1:T}).
\label{eq:approx1}
\end{align}
By using the linear system equations defining the evolution of the continuous hidden variables we find
\begin{align*}
&\av{(h_{t+1}-\hat h_{t+1}^{t,i,j,l})(h_t-\hat h_t^{t,i,l})\trans}_{p(h_{t+1}|s_t=i,s_{t+1}=j,c_t=l,v_{1:t})}=A^jP_t^{t,i,l}.
\end{align*}
Thus the joint covariance of $p(h_{t:t+1}|s_{t:t+1},c_t,v_{1:t})$ is given by
\begin{displaymath}
\left( \begin{array}{cc}
P_t^{t,i,l} &  P_t^{t,i,l}(A^j)\trans \\
A^jP_t^{t,i,l} & A^jP_t^{t,i,l}(A^j)\trans+\Sigma^j_H\\
\end{array} \right)\,.
\end{displaymath}
By using the formula of Gaussian conditioning we find the $p(h_t|h_{t+1},\sigma_{t},s_{t+1},v_{1:t})$ is Gaussian with mean and covariance given by
\begin{align*}
&\hat h_t^{t,\sigma_t}- \hat A^{\sigma_t,s_{t+1}}_t(h_{t+1}-A^{s_{t+1}}\hat h_t^{t,\sigma_t})\\
&P_t^{t,\sigma_t}- \hat A^{\sigma_t,s_{t+1}}_tA^{s_{t+1}}P_t^{t,\sigma_t}\, .
\end{align*}
where $\hat A^{\sigma_t,s_{t+1}}_t=P_t^{t,\sigma_t}(A^{s_{t+1}})\trans(A^{s_{t+1}}P_t^{t,\sigma_t}(A^{s_{t+1}})\trans+\Sigma^{s_{t+1}}_H)^{-1}$.
We can now define
\begin{align*}
h_t=\hat A^{\sigma_t,s_{t+1}}_th_{t+1}+\hat m^{\sigma_t,s_{t+1}}_t+\hat \eta^{\sigma_t,s_{t+1}}_t\, ,
\end{align*}
where
$m^{\sigma_t,s_{t+1}}_t=\hat h^{t,\sigma_t}_t-\hat A^{\sigma_t,s_{t+1}}_tA^{s_{t+1}}\hat h^{t,\sigma_t}_t$ and
$\hat\eta^{\sigma_t,s_{t+1}}_t={\cal N}(0,P_t^{t,\sigma_t}-\hat A^{\sigma_t,s_{t+1}}_tA^{s_{t+1}}P_t^{t,\sigma_t})$
which gives a Gaussian distribution for $p(h_t|\sigma_{t:t+1},v_{1:T})$ with mean and covariance
\begin{align*}
&\hat h_t^{T,\sigma_{t:t+1}}=\hat A^{\sigma_t,s_{t+1}}_t\hat h_{t+1}^{T,\sigma_{t+1}}+\hat m^{\sigma_t,s_{t+1}}_t\\
&P_t^{T,\sigma_{t:t+1}}=\hat A^{\sigma_t,s_{t+1}}_tP_{t+1}^{T,\sigma_{t+1}}(\hat A^{\sigma_t,s_{t+1}}_t)\trans+P_t^{t,\sigma_t}-\hat A^{\sigma_t,s_{t+1}}_tP_t^{t,\sigma_t}A^{s_{t+1}}\, .
\end{align*}
Therefore $p(h_t|\sigma_t,v_{1:T})$ is a mixture of Gaussians with mixture components $p(\sigma_{t+1}|\sigma_t,v_{1:T})$.

\paragraph{Computing $p(\sigma_{t+1}|\sigma_t,v_{1:T})$.}
We first observe that $p(\sigma_t|v_{1:T})$ can be computed recursively as $p(\sigma_t|v_{1:T})=\sum_{\sigma_{t+1}}p(\sigma_t|\sigma_{t+1},v_{1:T})p(\sigma_{t+1}|v_{1:T})$ where
\begin{align*}
&p(\sigma_t|\sigma_{t+1},v_{1:T})=\int_{h_{t+1}}p(\sigma_t|h_{t+1},\sigma_{t+1},v_{1:t},\cancel{v_{t+1:T}})p(h_{t+1}|\sigma_{t+1},v_{1:T})\, .
\end{align*}
The above integral, the average of the function $p(\sigma_t|h_{t+1},\sigma_{t+1},v_{1:t})$ wrt $p(h_{t+1}|\sigma_{t+1},v_{1:T})$, cannot be estimated in closed form. On the line of \citep{barber06ec}, we thus approximate it as follows
\begin{align}
&\hspace{-0cm}\av{p(\sigma_t|h_{t+1},\sigma_{t+1},v_{1:t})}_{p(h_{t+1}|\sigma_{t+1},v_{1:T})}\approx p(\sigma_t|h_{t+1},\sigma_{t+1},v_{1:t})|_{h_{t+1}=\hat h_{t+1}^{T,\sigma_{t+1}}}\, .
\label{eq:approx2}
\end{align}
The term $p(\sigma_{t+1}|\sigma_t,v_{1:T})$ can then be easily derived.
\vspace{-0.3cm}
\paragraph{Gaussian Collapsing}
\begin{align*}
&\hat h_t^{T,\sigma_t}=\sum_{\sigma_{t+1}}p(\sigma_{t+1}|\sigma_t,v_{1:T})\hat h_t^{T,\sigma_{t:t+1}}\\
&P_t^{T,\sigma_t}=\sum_{\sigma_{t+1}}p(\sigma_{t+1}|\sigma_t,v_{1:T})(P_t^{T,\sigma_{t:t+1}}+\hat h_t^{T,\sigma_{t:t+1}}(\hat h_t^{T,\sigma_{t:t+1}})\trans)-\hat h_t^{T,\sigma_t}(\hat h_t^{T,\sigma_t})\trans.
\end{align*}
Since collapsing is required only for $c_t=1$, the computational cost of the smoothing recursion is given by $O(TSd_{\max})$.
\paragraph{Cut of dependence when changing regime.}
\begin{align*}
p(h_t|\sigma_t,v_{1:T})
&=\delta_{c_t=1}\sum_{s_{t+1}}p(h_t|\sigma_t,\cancel{\sigma_{t+1}},v_{1:t},\cancel{v_{t+1:T}})\sum_{c_{t+1}}p(\sigma_{t+1}|\sigma_t,v_{1:T})\\
&+\delta_{c_t>1}p(h_t|\sigma_t,s_{t+1}=s_t,c_{t+1}=c_t-1,v_{1:T})\\
&=\delta_{c_t=1}p(h_t|\sigma_t,v_{1:t})\\
&+\delta_{c_t>1}\int_{h_{t+1}}p(h_t|h_{t+1},\sigma_t,s_{t+1}=s_t,\cancel{c_{t+1}=c_t-1},v_{1:t},\cancel{v_{t+1:T}})p(h_{t+1}|\sigma_{t:t+1},v_{1:T})
\end{align*}
From this recursion, it would look like approximations as (\ref{eq:approx1}) and (\ref{eq:approx1}) are required.
As we will see below, when using a modeling for $c_{1:T}$ as in \secref{sec:M1alt},
these will become exact and thus inference can be computed exactly.

\subsubsection{Increasing Count-Duration Variables}
Using a modeling for $c_{1:T}$ as in \secref{sec:M1alt} is beneficial over \secref{sec:SLGSSMM1st} when cutting dependence across regimes, in which case filtering becomes computationally less expensive and the smoothing approximations (\ref{eq:approx1}) and (\ref{eq:approx1}) become exact.

\subsubsection*{Filtering}
We can introduce cut of dependence when changing regime by adding a link from $c_t$ to $h_t$ as in \figref{fig:SLGSS} (c). The filtering recursion becomes
\begin{align*}
p(h_t|\sigma_t,v_{1:t})&=\delta_{c_t=1}\frac{p(v_t|h_t,s_t){\cal N}(h_t;\mu^{s_t},\Sigma^{s_t})}{p(v_t|s_t,c_t)}\\
&+\delta_{c_t>1}p(h_t|s_{t-1}=s_t,c_{t-1}=c_t-1,\sigma_t,v_{1:t})\overbrace{p(s_{t-1}=s_t,c_{t-1}=c_t-1|\sigma_t,v_{1:t})}^{1}\\
&=\delta_{c_t=1}\frac{p(v_t|h_t,s_t){\cal N}(h_t;\mu^{s_t},\Sigma^{s_t})}{p(v_t|s_t,c_t)}\\
&+\delta_{c_t>1}\frac{p(v_t|h_t,s_t)\int_{h_{t-1}}{\cal N}(h_t;A^{s_t}h_{t-1},\Sigma^{s_t}_H)p(h_{t-1}|s_{t-1}\!=\!s_t,c_{t-1}\!=\!c_t-1,\cancel{\sigma_t},v_{1:t-1})}{p(v_t|s_t,c_t)}\\
\end{align*}
From this recursion we see that $p(h_t|\sigma_t,v_{1:t})$ is a Gaussian (and therefore $p(h_t|v_{1:t})$ is a mixture of Gaussian with $Sd_{\max}$ components, as opposed to $TSd_{\max}$ components of \secref{sec:SLGSSMM1st}).

Notice that a standard change-point model without explicit duration distribution can be obtained by setting
$c_t=\{2,1\}$ with
\begin{align*}
p(h_t|h_{t-1},s_t,c_{t})=\begin{cases}
{\cal N}(h_t;\mu^{s_t},\Sigma^{s_t}) \hspace{0.57cm}	& \textrm{if } c_{t}=1\\
{\cal N}(h_t;A^{s_t}h_{t-1},\Sigma^{s_t}_H) 	& \textrm{if } c_{t}=2	
\end{cases}
\hspace{0.3cm}
p(c_t|s_{t-1},s_t)=\begin{cases}
    1& \textrm{if } s_{t-1}\neq s_t\\
	2& \textrm{if } s_{t-1}=s_t
\end{cases}
\end{align*}
Notice that, in this case, $p(h_t|s_t,c_t=1,v_{1:t})$ is Gaussian, whilst $p(h_t|s_t,c_t=2,v_{1:t})$ is a mixture of Gaussian with $t-1$ components (the value $c_t$ does not give information on the value of $c_{t-1}$ as in the previous case), and therefore $p(h_t|,v_{1:t})$ is a mixture of Gaussian with $St$ components.
\subsubsection*{Smoothing}

If we use a modeling for as in \secref{sec:M1alt}, we obtain
\begin{align*}
p(h_t|\sigma_t,v_{1:T})&=\sum_{\sigma_{t+1}}p(h_t|\sigma_{t:t+1},v_{1:T})p(\sigma_{t+1}|\sigma_t,v_{1:T})\\
&=\delta_{c_t\geq d_{\min}}\sum_{s_{t+1}}p(h_t|\sigma_t,\cancel{s_{t+1}},c_{t+1}=1,v_{1:t},\cancel{v_{t+1:T}})p(s_{t+1},c_{t+1}=1|\sigma_t,v_{1:T})\\
&+\delta_{c_t<d_{\max}}p(h_t|\sigma_t,s_{t+1}=s_t,c_{t+1}=c_t+1,v_{1:T})\overbrace{p(s_{t+1}=s_t,c_{t+1}=c_t+1|\sigma_t,v_{1:T})}^{1}\\
&=p(h_t|\sigma_t,v_{1:t})p(c_{t+1}=1|\sigma_t,v_{1:T})\\
&+\delta_{c_t<d_{\max}}\int_{h_{t+1}}p(h_t|h_{t+1},\sigma_t,s_{t+1}=s_t,\cancel{c_{t+1}=c_t+1},v_{1:t},\cancel{v_{t+1:T}})\\
&p(h_{t+1}|\sigma_t,s_{t+1}=s_t,c_{t+1}=c_t+1,v_{1:T})\overbrace{p(s_{t+1}=s_t,c_{t+1}=c_t+1|\sigma_t,v_{1:T})}^{1}
\end{align*}
Notice that, since $c_{t+1}=d>1, s_{t+1}=k$ implies $s_t=k,c_t=d-1$, we have $p(h_{t+1}|s_t=k,c_t=d-1,s_{t+1}=k,c_{t+1}=d,v_{1:T})=p(h_{t+1}|s_{t+1}=k,c_{t+1}=d,v_{1:T})$,
and therefore the approximation in \eqref{eq:approx1} becomes exact.

Notice that, since we established that $p(h_t|\sigma_t,v_{1:t})$ is Gaussian, at time $t$ $p(h_t|\sigma_t,v_{1:T})$ is a mixture of $T-t+1$
Gaussians.

\paragraph{Computing $p(\sigma_{t+1}|\sigma_t,v_{1:T})$.}
We first observe that $p(\sigma_t|v_{1:T})$ can be computed recursively as $p(\sigma_t|v_{1:T})=\sum_{\sigma_{t+1}}p(\sigma_t|\sigma_{t+1},v_{1:T})p(\sigma_{t+1}|v_{1:T})$ where
\begin{align*}
p(\sigma_t|v_{1:T})&=\sum_{\sigma_{t+1}}p(\sigma_t|\sigma_{t+1},v_{1:T})p(\sigma_{t+1}|v_{1:T})\\
&=\sum_{s_{t+1}}p(\sigma_t|\cancel{s_{t+1}},\cancel{c_{t+1}=1},v_{1:t},\cancel{v_{t+1:T}})p(s_{t+1},c_{t+1}=1|v_{1:T})\\
&+p(\sigma_t|s_{t+1}=s_t,c_{t+1}=c_t+1,v_{1:T})p(\sigma_{t+1}|v_{1:T})\\
&p(\sigma_t|v_{1:t})p(c_{t+1}=1|v_{1:T})+p(\sigma_{t+1}|v_{1:T})
\end{align*}
The above average $\av{p(\sigma_t|h_{t+1},\sigma_{t+1},v_{1:t})}_{p(h_{t+1}|\sigma_{t+1},v_{1:T})}$ cannot be estimated in closed form. We therefore approximate it by evaluation at the mean
\begin{align*}
&\hspace{-0cm}\av{p(\sigma_t|h_{t+1},\sigma_{t+1},v_{1:t})}_{p(h_{t+1}|\sigma_{t+1},v_{1:T})}\approx p(\sigma_t|h_{t+1},\sigma_{t+1},v_{1:t})|_{h_{t+1}=\hat h_{t+1}^{T,\sigma_{t+1}}}\, ,
\end{align*}
on the line of \citep{barber06ec}. The term $p(\sigma_{t+1}|\sigma_t,v_{1:T})$ can then be easily derived.

\vspace{-0.2cm}
\section{Conclusions}
\vspace{-0.3cm}
In this paper, we gave a unified and simple overview of many probabilistic models
used for modeling time-series data containing different underlying dynamical regimes. This common
framework enabled us to make connections among models that were not observed in the past,
and also to naturally introduce a variety of new models and inference routines.

The paper is therefore to serve as a tutorial for researchers wishing to gain an introduction into the subject, with the advantage that the material is presented under a common and consistent modelling framework. It is also intended to be of benefit to researchers working in the area, both by offering a different viewpoint to previous classical review papers and also by introducing several important extensions to current models.

\section*{Acknowledgments}
The author would like to thank the European Community for
supporting her research through a Marie Curie Intra European Fellowship.

\bibliographystyle{apalike}
\bibliography{UHMSM}
\end{document}